\documentclass[10pt,twocolumn,letterpaper]{article}

\usepackage{iccv}
\usepackage{times}
\usepackage{epsfig}
\usepackage{graphicx}
\usepackage{amsmath}
\usepackage{amssymb}

\usepackage[pagebackref=true,breaklinks=true,letterpaper=true,colorlinks,bookmarks=false]{hyperref}

\iccvfinalcopy %

\def\iccvPaperID{***} %

\ificcvfinal\fi

\usepackage[dvipsnames]{xcolor}
\usepackage{amsfonts}
\usepackage{bm}
\usepackage{amsmath}
\usepackage{mathtools}
\usepackage{multirow}
\usepackage{booktabs}
\usepackage{caption}
\usepackage{enumitem}
\usepackage{cleveref}
\usepackage{hyperref}

\newcommand{\acknowledgments}[1]{\noindent{\textbf{Acknowledgments. }}}
\renewcommand{\paragraph}[1]{{\vspace{1mm}\noindent \bf #1}.}
\newcommand{\paragraphnoper}[1]{{\vspace{1mm}\noindent \bf #1}}

\newcommand{\x}{\mathbf{x}}
\newcommand{\xt}{\x_t}
\newcommand{\xzero}{\x_0}

\newcommand{\pxt}{\hat{\x}_t}
\newcommand{\xtm}{\x_{t-1}}
\newcommand{\pxtm}{\hat{\x}_{t-1}}
\newcommand{\z}{\mathbf{z}}
\newcommand{\zt}{\z_t}
\newcommand{\zall}{\z_{1:T}}
\newcommand{\zleq}{\z_{\leq t}}
\newcommand{\zprev}{\z_{<t}}
\newcommand{\zpose}{\z^\text{pose}}
\newcommand{\trans}{\mathbf{r}}
\newcommand{\transvel}{\dot{\trans}}
\newcommand{\orient}{\Phi}
\newcommand{\orientvel}{\dot{\Phi}}
\newcommand{\pose}{\Theta}
\newcommand{\joints}{\mathbf{J}}

\newcommand{\predjointstime}{\hat{\mathbf{J}}_t}

\newcommand{\jointsvel}{\dot{\mathbf{J}}}
\newcommand{\mesh}{\mathbf{V}}
\newcommand{\meshtime}{\mathbf{V}_t}

\newcommand{\predmeshtime}{\hat{\mathbf{V}}_t}
\newcommand{\smpljoints}{\joints^{\text{SMPL}}}
\newcommand{\smpljointstime}{\joints^{\text{SMPL}}_t}
\newcommand{\predsmpljointstime}{\hat{\joints}^{\text{SMPL}}_t}
\newcommand{\shape}{\beta}
\newcommand{\RThree}{\mathbb{R}^3}
\newcommand{\schedsamp}{s_i}

\newcommand{\pnt}{\mathbf{p}}

\newcommand{\obszero}{\mathbf{y}_0}
\newcommand{\obst}{\mathbf{y}_t}
\newcommand{\obsall}{\mathbf{y}_{0:T}}

\newcommand{\ptheta}{p_\theta}
\newcommand{\decoder}{\ptheta(\xt | \zt, \xtm)}
\newcommand{\qphi}{q_\phi}
\newcommand{\encoder}{\qphi(\zt | \xt, \xtm)}

\newcommand{\prior}{\ptheta(\zt | \xtm)}
\newcommand{\DKL}{D_{\text{KL}}}

\newcommand{\normal}{\mathcal{N}}

\newcommand{\floor}{\mathbf{g}}

\newcommand{\loss}[1]{\mathcal{L}_{\text{#1}}}
\newcommand{\w}[1]{w_{\text{#1}}}
\newcommand{\enw}[1]{\lambda_{\text{#1}}}
\newcommand{\e}[1]{e^{#1}}

\newcommand{\energy}{{\mathcal{E}}}
\newcommand{\en}[1]{\energy_{\text{#1}}}
\newcommand{\ensub}[2]{\energy_{\text{#1}}^{\text{#2}}}

\DeclarePairedDelimiterX{\infdivx}[2]{(}{)}{%
  #1\;\delimsize\|\;#2%
}
\newcommand{\infdiv}{\DKL\infdivx}

\newcommand{\reals}{\mathbb{R}}

\crefname{appendix}{App.}{Appendix}
\crefname{section}{Sec.}{Section}
\crefname{equation}{Eq.}{Eq}
\crefname{figure}{Fig.}{Figure}
\crefname{table}{Tab.}{Table}

\newcommand{\enc}[1][encoder]{\textcolor{ForestGreen}{\textbf{#1}}\xspace}
\newcommand{\cpr}[1][conditional prior]{\textcolor{orange}{\textbf{#1}}\xspace}
\newcommand{\dec}[1][decoder]{\textcolor{RoyalBlue}{\textbf{#1}}\xspace}

\newcommand{\cdecoder}{\textcolor{RoyalBlue}{\ptheta(\xt | \zt, \xtm)}}
\newcommand{\cencoder}{\textcolor{ForestGreen}{\qphi(\zt | \xt, \xtm)}}
\newcommand{\cprior}{\textcolor{orange}{\ptheta(\zt | \xtm)}}

\newcommand{\methodname}{{{HuMoR}}\xspace}
\newcommand{\optname}{{{TestOpt}}\xspace}
\definecolor{contactcolor}{rgb}{0.80,0.20,0.20}
\definecolor{gndcolor2}{rgb}{0.3529412,0.4784314,0.4784314}
\definecolor{gndcolor1}{rgb}{0.254902,0.580392,0.580392}

\begin{document}

\title{\methodname: 3D \textbf{Hu}man \textbf{Mo}tion \textbf{Mo}del for \textbf{R}obust Pose Estimation}
\vspace{-5pt}
\author{Davis Rempe$^{1}$\qquad Tolga Birdal$^{1}$\qquad Aaron Hertzmann$^{2}$ \qquad Jimei Yang$^{2}$\\ \qquad Srinath Sridhar$^{3}$ \qquad Leonidas J. Guibas$^{1}$\\ \vspace{1mm}
\text{\normalsize $^1$Stanford University\qquad $^2$Adobe Research \qquad $^3$Brown University}\\
}

\twocolumn[{
\renewcommand\twocolumn[1][]{#1}%
\maketitle
\vspace{-0.485in}
\begin{center}
    \centering
    \includegraphics[width=\textwidth]{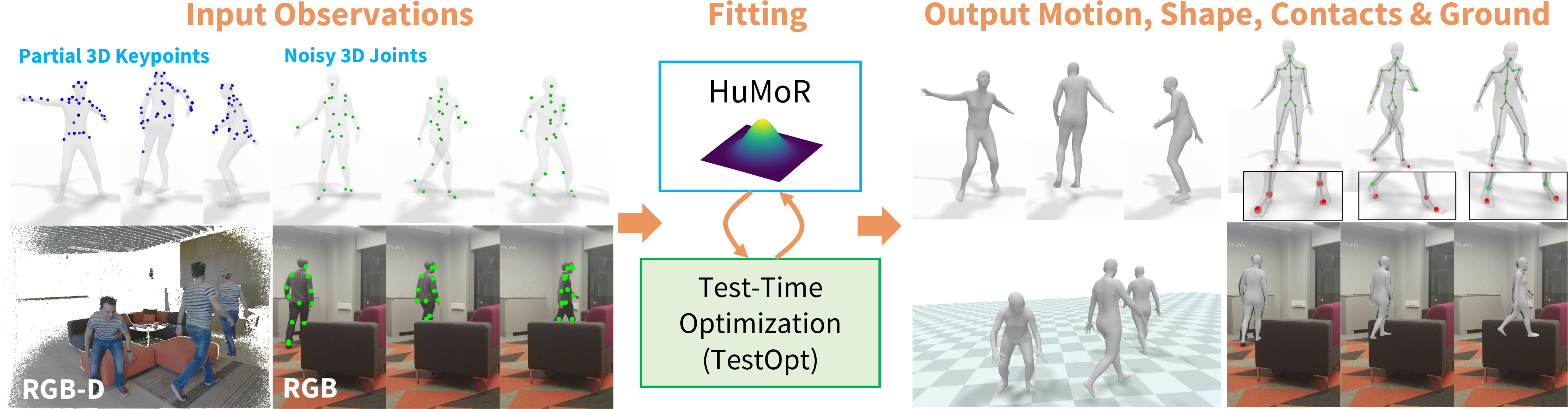}
    \captionof{figure}{\emph{Overview}. \methodname is a 3D \textbf{Hu}man \textbf{Mo}tion model for \textbf{R}obust estimation of temporal pose formulated as a conditional variational autoencoder. (Left) The proposed approach can operate on many input modalities and is designed to handle partial and noisy observations. (Middle/Right) A test-time optimization fits 3D motion and shape to an input sequence using \methodname as a prior; additional outputs include the ground and person-ground contacts (colored as \textcolor{gndcolor1}{ground} \textcolor{gndcolor2}{plane} and \textcolor{contactcolor}{contacts}).\vspace{-1mm}}
    \label{fig:teaser}
\end{center}%
}]
\maketitle
\ificcvfinal\thispagestyle{empty}\fi

\begin{abstract}
 We introduce \methodname: a 3D \textbf{Hu}man \textbf{Mo}tion Model for \textbf{R}obust Estimation of temporal pose and shape. Though substantial progress has been made in estimating 3D human motion and shape from dynamic observations, recovering plausible pose sequences in the presence of noise and occlusions remains a challenge. For this purpose, we propose an expressive generative model in the form of a conditional variational autoencoder, which learns a distribution of the change in pose at each step of a motion sequence.  Furthermore, we introduce a flexible optimization-based approach that leverages \methodname as a motion prior to robustly estimate plausible pose and shape from ambiguous observations. Through extensive evaluations, we demonstrate that our model generalizes to diverse motions and body shapes after training on a large motion capture dataset, and enables motion reconstruction from multiple input modalities including 3D keypoints and RGB(-D) videos. 
 See the project page at \href{https://geometry.stanford.edu/projects/humor/}{\texttt{geometry.stanford.edu/projects/humor}}.
\end{abstract}
\section{Introduction}
As humans, we are constantly moving in, interacting with, and manipulating the world around us.
Thus, applications such as action recognition~\cite{varol2019synthetic,varol2017learning} or 
holistic dynamic indoor scene understanding~\cite{chen2019holistic++} require accurate perception of 3D human pose, shape, motion, contacts, and interaction.
Extensive previous work has focused on estimating 2D or 3D human pose~\cite{cao2019openpose,mehta2020xnect,mehta2017vnect}, shape~\cite{pavlakos2019smplx,habermann2020deepcap,Saito_2019_ICCV}, and motion~\cite{kocabas2019vibe} from videos.
These are challenging problems due to the large space of articulations, body shape, and appearance variations. Even the best methods struggle to accurately capture a wide variety of motions from varying input modalities, producing noisy or overly-smoothed motions (especially at ground contact, \ie,~footskate), and struggle with occlusions (\eg,~walking behind a couch as in~\cref{fig:teaser}).

We focus on the problem of building a robust human motion model that can address these challenges.
To date, most motion models directly represent sequences of likely poses --- \eg, in PCA space \cite{DBLP:conf/nips/OrmoneitSBH00,Urtasun:2006:CVIU,Sidenbladh:2000:STH} or via future-predicting autoregressive processes~\cite{Taylor:2007:MHM,Urtasun:2006:CVPR,Pavlovic:2000:LSL}.
However, purely pose-based predictions either make modeling environment interactions and generalization beyond training poses difficult, or quickly diverge from the space of realistic motions. 
On the other hand, explicit physical dynamics models \cite{RempeContactDynamics2020,li2019estimating,shimada2020physcap,Peng:2018:SFV,Brubaker:2009:ECD,Brubaker:2010:PBP} are resource intensive and require knowledge of unobservable physical quantities.
While generative models potentially offer the required flexibility, building an \emph{expressive},  \emph{generalizable} and \emph{robust} model for \emph{realistic} 3D human motions remains an open problem.

To address this, we introduce a learned, autoregressive, generative model that captures the \emph{dynamics} of 3D human motion, \ie, how pose changes over time. 
Rather than describing likely poses, the \textbf{Hu}man \textbf{Mo}tion Model for \textbf{R}obust Estimation (\methodname) models a probability distribution of possible \emph{pose transitions}, formulated as a conditional variational autoencoder \cite{sohn2015CVAE}. Though not explicitly physics-based, its components correspond to a physical model: the latent space can be interpreted as generalized forces, which are inputs to a dynamics model with numerical integration (the decoder). Moreover, ground contacts are explicitly predicted and used to constrain pose estimation at test time.

After training on the large AMASS motion capture dataset \cite{mahmood2019amass}, we use \methodname as a \emph{motion prior} at test time for 3D human perception from noisy and partial observations across different input modalities such as \mbox{RGB(-D)} video and 2D or 3D joint sequences, as illustrated in~\cref{fig:teaser} (left).
In particular, we introduce a robust test-time optimization strategy which interacts with \methodname to estimate the parameters of \emph{3D motion}, \emph{body shape}, the \emph{ground plane}, and \emph{contact points} as shown in~\cref{fig:teaser} (middle/right). 
This interaction happens in two ways: (i) by parameterizing the motion in the latent space of \methodname,
and (ii) using \methodname priors in order to regularize the optimization towards the space of plausible motions.

Comprehensive evaluations reveal that our method surpasses the state-of-the-art on a variety of visual inputs in terms of accuracy and physical plausibility of motions under partial and severe occlusions. We further demonstrate that our motion model generalizes to diverse motions and body shapes on common generative tasks like sampling and future prediction. 
In a nutshell, our contributions are:
\begin{itemize}[leftmargin=\parindent]
\item \methodname, a generative 3D human motion prior modeled by a novel conditional VAE which enables expressive and general motion reconstruction and generation,
\item A subsequent robust test-time optimization approach that uses \methodname as a strong motion prior jointly solving for pose, body shape, and ground plane / contacts, %
\item The capability to operate on a variety of inputs, such as \mbox{RGB(-D)} video and 2D/3D joint position sequences, to yield accurate and plausible motions and contacts, exemplified through extensive evaluations.
\end{itemize}

Our work, more generally, suggests that neural nets for dynamics problems can benefit from architectures that model transitions, allowing control structures that emulate classical physical formulations.
\section{Related Work}
Much progress has been made on building methods to recover 3D joint locations~\cite{pavllo20193d,mehta2017vnect,mehta2020xnect} or parameterized 3D pose and shape (\ie, SMPL~\cite{loper2015SMPL}) from observations~\cite{varol18_bodynet}.
We focus primarily on motion and shape estimation. %

\paragraph{Learning-Based Estimation}
Deep learning approaches have shown success in regressing 3D shape and pose from a single image~\cite{SPIN:ICCV:2019,hmrKanazawa17,pavlakos2018learning,guler2018densepose,guler2019holopose,zanfir2020weakly,choutas2020monocular}. This has led to developments in predicting \emph{motion} (pose sequences) and shape directly from RGB video~\cite{kanazawa2019learning,zhang2019predicting,shi2020motionet,sun2019human,doersch2019sim2real}. Most recently, VIBE~\cite{kocabas2019vibe} uses adversarial training to encourage plausible outputs from a conditional recurrent motion generator. MEVA~\cite{luo20203d} maps a fixed-length image sequence to the latent space of a pre-trained motion autoencoder. These methods are fast and produce accurate root-relative joint positions for video, but motion is globally inconsistent and they struggle to generalize, \eg, under severe occlusions.
Other works have addressed occlusions
but only on static images~\cite{biggs20203d,zhang2020object,rockwell2020full,georgakis2020hierarchical,KocabasPARE2021}.
Our approach resolves difficult occlusions in video and other modalities by producing plausible and expressive motions with HuMoR.

\paragraph{Optimization-Based Estimation}
 One may directly optimize to more accurately fit to observations (images or 2D pose estimators~\cite{cao2019openpose}) using human body models~\cite{ganapathi2010real,baak2013data,bogo2016smplify}. SMPLify~\cite{bogo2016smplify} uses the SMPL model~\cite{loper2015SMPL} to fit pose and shape parameters to 2D keypoints in an image using priors on pose and shape. 
 Later works consider body silhouettes~\cite{lassner2017unite} and use a learned variational pose prior~\cite{pavlakos2019smplx}. Optimization for motion sequences has been explored by several works~\cite{arnab2019exploiting,huang2017towards,liu20204d,zanfir2018monocular,xiang2019monocular} which apply simple smoothness priors over time. These produce reasonable estimates when the person is fully visible, but with unrealistic dynamics, \eg, overly smooth motions and footskate.
 
 Some works employ human-environment interaction and contact constraints to improve shape and pose estimation~\cite{hassan2019prox,liu20204d,hassan2020POSA} by assuming scene geometry is given. %
 iMapper ~\cite{monszpart2019iMapper} recovers both 3D joints and a primitive scene representation from RGB video based on interactions by motion retrieval,
 which may differ from observations.
 In contrast, our approach optimizes for pose and shape by using an expressive generative model that produces more natural motions than prior work with realistic ground contact.

\paragraph{Human Motion Models}
Early sophisticated motion models for pose tracking used a variety of approaches, including mixtures-of-Gaussians \cite{DBLP:conf/nips/HoweLF99}, linear embeddings of periodic motion \cite{DBLP:conf/nips/OrmoneitSBH00,Urtasun:2006:CVIU,Sidenbladh:2000:STH}, nonlinear embeddings \cite{Elgammal:2004:SSC}, and nonlinear autoregressive models \cite{Taylor:2007:MHM,Wang:2005:GPD,Urtasun:2006:CVPR,Pavlovic:2000:LSL}. These methods operate in pose space, and are limited to specific motions. 
Models based on physics can potentially generalize more accurately \cite{RempeContactDynamics2020,li2019estimating,shimada2020physcap,Peng:2018:SFV,Brubaker:2009:ECD,Brubaker:2010:PBP,yuan2021simpoe}, while also estimating global pose and environmental interactions.
However, general-purpose physics-based models are difficult to learn, computationally intensive at test-time, and often assume full-body visibility to detect contacts~\cite{RempeContactDynamics2020,li2019estimating,shimada2020physcap}.

Many motion models have been learned for computer animation \cite{Brand:2000:SM,Kovar:2002:MG,Rose:1998:VAA,Li:2002:MTA,Liu:2005:LPB,holden2017phase,starke2019neural} including recent recurrent and autoregressive models~\cite{habibie2017recurrent,ghorbani2020probabilistic,henter2020moglow,yang2021real,ling2020MVAE}. These often focus on visual fidelity for a small set of characters and periodic locomotions.
Some have explored generating more general motion and body shapes~\cite{zhang2020mojo,pavllo2019modeling,aksan2019structured,Corona_2020_CVPR}, but in the context of short-term future prediction.
\methodname is most similar to Motion VAE~\cite{ling2020MVAE}, however we make crucial contributions to enable generalization to unseen, non-periodic motions on novel body shapes.

\begin{figure*}
\begin{center}
\includegraphics[width=1.0\textwidth]{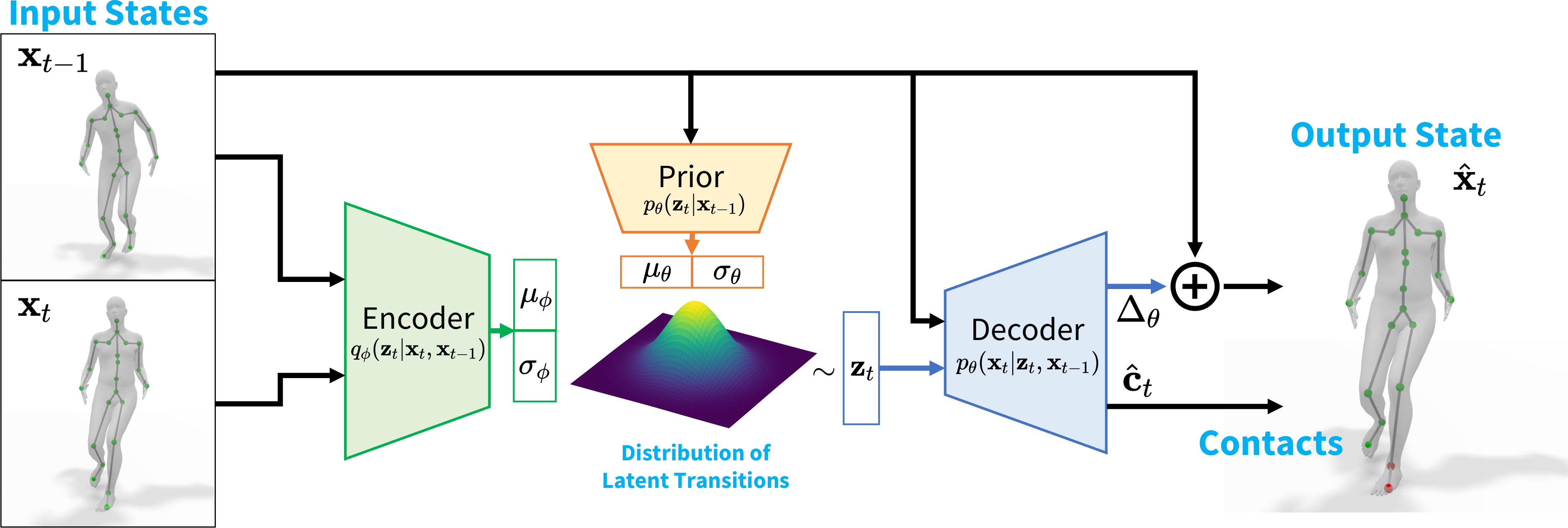}
\end{center}
   \caption{\emph{HuMoR CVAE Architecture}. During training, given the previous state $\xtm$ and ground truth current state $\xt$, the model reconstructs $\pxt$ by sampling from the \enc distribution. At test time we can (i) \emph{generate} the next state from $\xtm$ by sampling from the \cpr[prior]{} distribution and \dec[decoding], (ii) \emph{infer} a latent transition $\zt$ with the \enc, or (iii) evaluate the \emph{likelihood} of a given $\zt$ with the \cpr.
   }
\label{fig:arch}\vspace{-3mm}
\end{figure*}
\section{HuMoR: 3D Human Dynamics Model}
\label{sec:method:prior}

The goal of our work is to build an \emph{expressive} and \emph{generalizable} generative model of 3D human motion learned from real human motions, and to show that this can be used for robust test-time optimization (\optname) of pose and shape.
In this section, we first describe the model, \methodname. %

\paragraph{State Representation}
We represent the state of a moving person as a matrix $\x$ composed of a root translation $\trans \in \RThree$, root orientation $\orient \in \RThree$ in axis-angle form, body pose joint angles $\pose \in \reals^{3\times 21}$ and joint positions  $\joints\in\reals^{3\times 22}$:
\begin{align}
\label{eqn:staterep}
    \x = [ \quad \trans \quad \transvel \quad \orient \quad \orientvel \quad \pose \quad \joints \quad \jointsvel \quad ],
\end{align}
where $\transvel, \orientvel$ and $\jointsvel$ denote the root and joint velocities, respectively, giving $\x \in \reals^{3\times 69}$.
Part of the state, $(\trans,\orient,\pose)$, parameterizes the SMPL body model~\cite{loper2015SMPL,smplh:SIGGRAPHASIA:2017} which is a differentiable function $M(\trans,\orient,\pose,\shape)$ that maps to body mesh vertices $\mesh \in \reals^{3\times 6890}$ and joints $\smpljoints \in \reals^{3\times 22}$ given shape parameters $\shape \in \reals^{16}$. Our over-parameterization allows for two ways to recover the joints: (i) explicitly from $\joints$, (ii) implicitly through the SMPL map $M(\cdot)$.

\paragraph{Latent Variable Dynamics Model}
We are interested in modeling the probability of a time sequence of states
\begin{equation}
    \ptheta ( \x_0, \x_1, \dots, \x_T ) = \ptheta(\x_0) \prod_{t=1}^{T} \ptheta( \xt | \xtm)\,,
\end{equation}
where each state is assumed to be dependent on only the previous one and $\theta$ are learned parameters. Then $\ptheta(\xt | \xtm)$ must capture the \emph{plausibility} of a transition. %

We propose a \textbf{conditional variational autoencoder (CVAE)} which formulates the motion $p_\theta(\xt|\xtm)$ as a latent variable model as shown in~\cref{fig:arch}.
Following the original CVAE derivation~\cite{sohn2015CVAE}, our model contains two main components.
First, conditioned on the previous state $\xtm$, the distribution over possible latent variables $\zt\in\reals^{48}$ is described by a learned \cpr: 
\begin{equation}
\prior = \normal (\zt; \mu_\theta(\xtm), \sigma_\theta(\xtm))\,, 
\end{equation}
which parameterizes a Gaussian distribution with diagonal covariance via a neural network.
Intuitively, the latent variable $\zt$ represents the transition to $\xt$ and should therefore have different distributions given different $\xtm$. For example, an idle person has a large variation of possible next states while a person in midair is on a nearly deterministic trajectory. Learning the conditional prior significantly improves the ability of the CVAE to generalize to diverse motions and empirically stabilizes both training and \optname.

Second, conditioned on $\zt$ and $\xtm$, the \dec produces two outputs, $\Delta_{\theta}$ and $\mathbf{c}_t$.
The \emph{change in state} $\Delta_{\theta}$ defines the output distribution $\decoder$ through
\begin{equation}
\xt = \xtm +\Delta_\theta(\zt, \xtm) + \boldsymbol{\eta}, \qquad \boldsymbol{\eta} \sim \normal (\mathbf{0}, \mathbf{I}).
\label{eq:decoder}
\end{equation}
We find the additive update $\Delta_{\theta}$ improves predictive accuracy compared to direct next-step prediction.
The person-ground contact $\mathbf{c}_t$ is the probability that each of 8 body joints (\emph{left and right toes}, \emph{heels}, \emph{knees}, and \emph{hands}) is in contact with the ground at time $t$. Contacts are \emph{not} part of the input to the conditional prior, only an output of the decoder.
The contacts enable environmental constraints in \optname. %

The complete probability model for a transition is then:
\begin{align}
\label{eqn:lvm}
    \ptheta(\xt | \xtm) &= \int_{\zt} \ptheta(\zt | \xtm) \ptheta(\xt | \zt, \xtm).
\end{align}
Given an initial state $\x_0$, one can sample a motion sequence by alternating between sampling $\zt \sim \ptheta(\zt | \xtm)$ and sampling
$\xt \sim \ptheta(\xt | \zt, \xtm)$, from $t=1$ to $T$.
This model parallels a conventional stochastic physical model. The conditional prior can be seen as a controller, producing ``forces'' $\zt$ as a function of state $\xtm$, while the decoder acts like a combined physical dynamics model and Euler integrator of generalized position and velocity in \cref{eq:decoder}. %

In addition to this nice physical interpretation, our model is motivated by 
Motion VAE (MVAE)~\cite{ling2020MVAE}, which has recently shown promising results for single-character locomotion animation, 
also using a VAE for $\ptheta(\xt | \xtm)$. However, we find that directly applying MVAE for estimation does not give good results (\cref{sec:exp}). We overcome this by additionally learning a \emph{conditional} prior, modeling the \emph{change in state} and contacts, and encouraging \emph{consistency} between joint position and angle predictions (\cref{sec:training}).

\paragraph{Rollout}
We use our model to define a deterministic \emph{rollout function}, which is key to \optname.  Given an initial state $\x_0$ and a sequence of latent transitions $\zall$, we define a function $\x_T = f(\x_0, \zall)$ that deterministically maps the motion ``parameters'' ($\x_0,\zall)$ to the resulting state at time $T$. This is done through autoregressive rollout which decodes and integrates $\xt = \xtm + \Delta_\theta(\zt, \xtm)$ at each timestep.

\paragraph{Initial State GMM} We model $\ptheta (\x_0)$ with a Gaussian mixture model (GMM) containing $K=12$ components with weights $\gamma^i$, so that $\ptheta (\xzero) = \sum_{i=1}^K \gamma^i \normal (\x_0; \mu_\theta^i, \sigma_\theta^i)$.

\subsection{Training} 
\label{sec:training}
\noindent Our CVAE is trained using pairs of ($\xtm$, $\xt$). We consider the usual variational lower bound:
\begin{align}
\label{eqn:elbo}
\log \ptheta(\xt | &\xtm) \geq \mathbb{E}_{\qphi}[ \log \cdecoder ] \nonumber \\
&- \infdiv{\cencoder}{\cprior}.
\end{align}
The expectation term measures the reconstruction error of the \dec. The \enc, \ie approximate posterior, is introduced for training and parameterizes a Gaussian distribution $\encoder = \normal (\zt; \mu_\phi (\xt, \xtm), \sigma_\phi (\xt, \xtm))$. %
The KL divergence $\infdiv{\cdot}{\cdot}$ regularizes its output to be near the \cpr[prior]{}. 
Therefore, we seek the parameters $(\theta,\phi)$ that minimize the loss function
\begin{align}
\label{eqn:L} 
\loss{rec} + \w{KL}\loss{KL} + \loss{reg}
\end{align}
over all training pairs in our dataset, where $\loss{rec} + \w{KL}\loss{KL}$ is the lower bound in~\cref{eqn:elbo} with weight $\w{KL}$, and $\loss{reg}$ contains additional regularizers.

For a single training pair ($\xtm$, $\xt$), the reconstruction loss is computed as $\loss{rec} = || \xt - \pxt ||^2$ from the decoder output $\pxt = \xtm + \Delta_\theta(\zt, \xtm)$ with $\zt \sim \encoder$. Gradients are backpropagated through this sample using the reparameterization trick~\cite{kingma2013auto}.
The regularization loss contains two terms: $\loss{reg} = \loss{SMPL} + \w{contact}\loss{contact}$.
The SMPL term $\loss{SMPL} = \loss{joint} + \loss{vtx} + \loss{consist}$ %
uses the output of the body model with the estimated parameters and ground truth shape $[\predsmpljointstime, \predmeshtime] = M(\hat{\trans}_t,\hat{\orient}_t,\hat{\pose}_t,\shape)$:
\begin{gather}
    \loss{joint} = || \smpljointstime - \predsmpljointstime ||^2 \\
    \loss{vtx} = || \meshtime - \predmeshtime ||^2 \qquad
    \loss{consist} = ||  \predjointstime -  \predsmpljointstime||^2.
\end{gather}
The loss $\loss{consist}$ encourages consistency between regressed joints and those of the body model. The contact loss $\loss{contact} = \loss{BCE} + \loss{vel}$ contains two terms. The first supervises ground contact classification with a typical binary cross entropy; the second regularizes joint velocities to be consistent with contacts $\loss{vel} = \sum_j \hat{c}^j_t || \hat{\mathbf{v}}_t ||^2$ with $\hat{\mathbf{v}}_t \in \hat{\dot{\joints}}_t$ and $\hat{c}^j_t \in \hat{\mathbf{c}}_t$ the predicted probability that joint $j$ is in ground contact. We set $\w{contact} = 0.01$ and $\w{KL} = 4\e{-4}$.

The initial state GMM is trained separately with expectation-maximization on the same dataset used to train the CVAE.  %

\paragraph{Implementation Details}
To ease learning and improve generalization, our model operates in an aligned canonical coordinate frame at each step. All networks are 4 or 5 layer MLPs with ReLU activations and group normalization~\cite{wu2018group}. To combat posterior collapse~\cite{lucas2019collapse,ling2020MVAE,sohn2015CVAE}, we linearly anneal $\w{KL}$ during training~\cite{bowman2015generating}. Following~\cite{ling2020MVAE}, we also use scheduled sampling~\cite{bengio2015scheduled} to enable long-term generation by making the model robust to its own errors. Additional details are available in the supplementary.

\section{Test-time Motion Optimization}
\label{sec:optim}
We next use the space of motion learned by \methodname as a \emph{prior} in \optname to recover pose and shape from noisy and partial observations while ensuring plausibility. %

\subsection{Optimization Variables}
Given a sequence of observations $\obsall$, either as 2D/3D joints, 3D point clouds, or 3D keypoints, we seek the shape $\shape$ and a sequence of SMPL pose parameters $(\trans_{0:T},\orient_{0:T},\pose_{0:T})$ which describe the underlying motion being observed. We parameterize the optimized motion \textbf{using our CVAE} by the initial state $\xzero$ and a sequence of latent transitions $\zall$. Then at $T$ (and any intermediate steps) $\x_T = f(\x_0, \zall)$ is determined through model \emph{rollout} using the \dec as previously detailed.
Compared to directly optimizing SMPL~\cite{arnab2019exploiting,bogo2016smplify,huang2017towards}, this motion representation naturally encourages plausibility and is compact in the number of variables.
To obtain the transformation between the canonical coordinate frame in which our CVAE is trained and the observation frame used for optimization, we additionally optimize the ground plane of the scene $\floor \in \RThree$. %
All together, we simultaneously optimize initial state $\xzero$, a sequence of latent variables $\zall$, ground $\floor$, and shape $\shape$. We assume a static camera with known intrinsics.

\subsection{Objective \& Optimization}
The optimization objective can be formulated as a maximum a-posteriori (MAP) estimate (see supplementary), which seeks a motion that is plausible under our generative model while closely matching observations:
\begin{equation}
\label{eqn:objectives}
      \min_{\xzero,\zall,\floor,\shape} \, \en{mot} + \en{data} + \en{reg}.
\end{equation}
We next detail each of these terms which are the motion prior, data, and regularization energies. In the following, $\lambda$ are weights to determine the contribution of each term.

\paragraph{Motion Prior $\en{mot}$}
This energy measures the likelihood of the latent transitions $\zall$ and initial state $\xzero$ under the \methodname CVAE and GMM. It is $\en{mot}=\en{CVAE}+\en{init}$ where
\begin{align}
    \en{CVAE} &= - \enw{CVAE}\sum_{t=1}^T \log \normal (\zt; \mu_\theta(\xtm), \sigma_\theta(\xtm))\nonumber \\
    \en{init} &= - \enw{init}\log \sum_{i=1}^K \gamma^i \normal (\xzero; \mu_\theta^i, \sigma_\theta^i).
\end{align}
$\en{CVAE}$ uses the learned \cpr and $\en{init}$ uses the initial state GMM. 

\paragraph{Data Term $\en{data}$}
This term is the \emph{only} modality-dependent component of our approach, requiring different losses for different inputs: 3D joints, 2D joints, and 3D point clouds. %
All data losses operate on SMPL joints or mesh vertices obtained through the body model $[\smpljoints_t, \mesh_t] = M(\trans_t,\orient_t,\pose_t,\shape)$ using the current shape $\shape$ along with the SMPL parameters $(\trans_t,\orient_t,\pose_t)$ contained in $\xt=f(\xzero, \z_{1:t})$ transformed from the canonical to observation (\ie camera) frame.
In the simplest case, the observations $\obst$ are 3D joint positions (or keypoints with known correspondences) and our energy is
\begin{equation}
\en{data}\triangleq\ensub{data}{3D} = \enw{data}\sum_{t=0}^{T} \sum_{j=1}^{J} || \pnt_t^j - \obst^j ||^2
\end{equation}
with $\pnt_t^j \in \smpljoints_t$. %
For 2D joint positions, each with a detection confidence $\sigma_t^j$, we use a re-projection loss
\begin{equation}
\en{data}\triangleq\ensub{data}{2D} =\enw{data} \sum_{t=0}^{T} \sum_{j=1}^{J} \sigma_t^j \rho( \Pi (\pnt_t^j) - \obst^j )
\end{equation}
with $\rho$ the robust Geman-McClure function~\cite{bogo2016smplify,geman1987gmef} and $\Pi$ the pinhole projection. If an estimated person segmentation mask is available, it is used to ignore spurious 2D joints. Finally, if $\obst$ is a 3D point cloud obtained from a depth map roughly masked around the person of interest, we use the mesh vertices to compute
\begin{equation}
\en{data}\triangleq\ensub{data}{PC3D} = \enw{data} \sum_{t=0}^{T} \sum_{i=1}^{N_t} \w{bs} \min_{\pnt_t \in \mesh_t} ||  \pnt_t - \obst^i  ||^2 %
\end{equation}
where %
$\w{bs}$ is a robust bisquare weight~\cite{beaton1974fitting} computed based on the Chamfer distance term.

\paragraph{Regularizers $\en{reg}$}
The additional regularization consists of four terms $\en{reg} = \en{skel} + \en{env} + \en{gnd} + \en{shape}$. The first two terms encourage rolled-out motions from the CVAE to be plausible even when the initial state $\x_0$ is far from the optimum (\ie early in optimization). The skeleton consistency term uses the joints $\joints_t$ directly \emph{predicted} by the decoder during rollout along with the SMPL joints:
\begin{equation}
    \en{skel} = \sum_{t=1}^{T}\Big( \enw{c}\sum_{j=1}^{J} || \pnt_t^j - \pnt_t^{j,\text{pred}}  ||^2 + \enw{b}\sum_{i=1}^{B} (l_t^i - l_{t-1}^i )^2\Big)\nonumber
\end{equation}
with $\pnt_t^{j} \in \smpljoints_t$ and $\pnt_t^{j,\text{pred}} \in \joints_t$.
The second summation uses bone lengths $l$ computed from $\joints_t$ at each step. The second regularizer $\en{env}$ ensures consistency between predicted CVAE contacts, the motion, and the environment:
\begin{equation}
    \en{env} = \sum_{t=1}^{T} \sum_{j=1}^{J} \enw{cv} c_t^j || \pnt_t^j - \pnt_{t-1}^j  ||^2 + \enw{ch} c_t^j \max(|\pnt_{z,t}^j| - \delta, 0)\nonumber
\end{equation}
where $\pnt_t^j \in \smpljoints_t$ and $c_t^j$ is the contact probability output from the model for joint $j$. The contact height term weighted by $\enw{ch}$ ensures the $z$-component of contacting joints are within $\delta$ of the floor in the canonical frame.

The final two regularizers are priors on the ground and shape. We assume the ground should stay close to initialization $\en{gnd} = \enw{gnd}|| \floor - \floor^\text{init} ||^2$. Finally, $\shape$ should stay near the neutral zero vector similar to~\cite{hassan2019prox,pavlakos2019smplx}: $\en{shape} = \enw{shape} || \shape ||^2$.

\paragraph{Initialization \& Optimization}
We initialize the temporal SMPL parameters $\trans_{0:T},\orient_{0:T},\pose_{0:T}$ and shape $\shape$ with an initialization optimization using $\en{data}$ and $\en{shape}$ along with two additional regularization terms. $\en{pose} = \sum_{t} || \zpose_t ||^2$ is a pose prior where $\zpose_t \in \mathbb{R}^{32}$ is the body joint angles represented in the latent space of the VPoser model~\cite{pavlakos2019smplx,hassan2019prox}. The smoothness term $\en{smooth} = \sum_{t=1}^{T} \sum_{j=1}^{J} || \pnt_t^j - \pnt_{t-1}^j  ||^2$ with $\pnt_t^j \in \smpljoints_t$ smooths 3D joint positions over time.
Afterwards, the initial latent sequence $\zall^\text{init}$ is computed through inference with the CVAE \enc.
Our optimization is implemented in PyTorch~\cite{paszke2017automatic} using L-BFGS and \emph{autograd}; with batching, a $3s$ RGB video takes about 5.5 $min$ to fit. We provide further details in the supplementary material.

\section{Experimental Results}\label{sec:exp}
\begin{table}
\setlength{\tabcolsep}{5pt}
\begin{center}
\begin{tabular}{l|ccc| c }
& \multicolumn{3}{c|}{Future Prediction} & Diversity \\
\textbf{Model} & \textbf{Contact} $\uparrow$ & \textbf{ADE} $\downarrow$ & \textbf{FDE} $\downarrow$ & \textbf{APD}  	$\uparrow$ \\
\hline
MVAE~\cite{ling2020MVAE} & - & 25.8 & 50.6 & 85.4 \\
\methodname & 0.88 & \textbf{21.5} & \textbf{42.1} & 94.9 \\
\methodname (Qual) & 0.88 & 22.0 & 46.3 & \textbf{100.0} 
\end{tabular}
\end{center}
\caption{(Left) Future prediction accuracy for $2s$ AMASS sequences. Contact classification accuracy, average displacement error ($cm$), and final displacement error ($cm$) are reported. (Right) Sampling diversity over $5s$ rollouts measured by average pairwise distance ($cm$).\vspace{-3mm}}
\label{table:motionprior}
\end{table}

We evaluate \methodname on (i) generative sampling tasks and (ii) as a prior in \optname to estimate motion from 3D and RGB(-D) inputs. We encourage viewing the \href{https://geometry.stanford.edu/projects/humor/supp.html}{\textbf{supplementary videos}} to appreciate the qualitative improvement of our approach. Additional dataset and experiment details are available in the supplementary document.

\subsection{Datasets}
\paragraphnoper{AMASS~\cite{mahmood2019amass}} 
is a large motion capture database containing diverse motions and body shapes on the SMPL body model. We sub-sample the dataset to 30 Hz and use the recommended training split to train the CVAE and initial state GMM in \methodname. We evaluate on the held out Transitions and HumanEva~\cite{sigal2010humaneva} subsets (\cref{expt:gen} and~\ref{expt:estim3d}).

\paragraphnoper{i3DB~\cite{monszpart2019iMapper}} contains RGB videos of person-scene interactions involving medium to heavy occlusions. It provides annotated 3D joint positions and a primitive 3D scene reconstruction which we use to fit a ground plane for computing plausibility metrics. We run off-the-shelf 2D pose estimation~\cite{cao2019openpose}, person segmentation~\cite{chen2017rethinking}, and plane detection~\cite{liu2019planercnn} models to obtain inputs for our optimization.

\paragraphnoper{PROX~\cite{hassan2019prox}} contains RGB-D videos of people interacting with indoor environments. We use a subset of the qualitative data to evaluate plausibility metrics using a floor plane fit to the provided ground truth scene mesh. We obtain 2D pose, person masks, and ground plane initialization in the same way as done for i3DB.

\subsection{Baselines and Evaluation Metrics}
\paragraph{Motion Prior Baselines} 
We ablate the proposed CVAE to analyze its core components: \emph{No Delta} directly predicts the next state from the decoder rather than the change in state, \emph{No Contacts} does not classify ground contacts, \emph{No $\loss{SMPL}$} does not use SMPL regularization in training, and \emph{Standard Prior} uses $\normal (\mathbf{0}, \mathbf{I})$ rather than our learned \cpr. All of these ablated together recovers \emph{MVAE}~\cite{ling2020MVAE}.

\paragraph{Motion Estimation Baselines}
\emph{VPoser-t} is the initialization phase of our optimization. It uses VPoser~\cite{pavlakos2019smplx} and 3D joint smoothing similar to previous works~\cite{arnab2019exploiting,huang2017towards,zanfir2018monocular}. \emph{PROX-(RGB/D)}~\cite{hassan2019prox} are optimization-based methods which operate on individual frames of RGB and RGB-D videos, respectively. Both assume the full scene mesh is given to enforce contact and penetration constraints. \emph{VIBE}~\cite{kocabas2019vibe} is a recent learned method to recover shape and pose from video. %

\paragraph{Error Metrics}
3D positional errors are measured on joints, keypoints, or mesh vertices (\textbf{Vtx}) and compute \emph{global} mean per-point position error unless otherwise specified. We report positional errors for all (\textbf{All}), occluded (\textbf{Occ}), and visible (\textbf{Vis}) observations separately. Finally, we report binary classification accuracy of the 8 person-ground contacts (\textbf{Contact}) predicted by \methodname.

\paragraph{Plausibility Metrics}
We use additional metrics to measure qualitative motion characteristics that joint errors cannot capture.
Smoothness is evaluated by mean per-joint accelerations (\textbf{Accel}) ~\cite{kanazawa2019learning}. 
Another important indicator of plausibility is ground penetration~\cite{RempeContactDynamics2020}. We use the true ground plane to compute the frequency (\textbf{Freq}) of foot-floor penetrations: the fraction of frames for both the left and right toe joints that penetrate more than a threshold. We measure frequency at 0, 3, 6, 9, 12, and $15~cm$ thresholds and report the mean. We also report mean penetration distance (\textbf{Dist}), where non-penetrating frames contribute a distance of 0 to make values comparable across differing frequencies.

\begin{table*}
\begin{center}
\setlength{\tabcolsep}{8.25pt}
\begin{tabular}{l c | c c c  c  c  c | c  c c}
 & & \multicolumn{3}{c}{Positional Error}  & Joints & Mesh &  & & \multicolumn{2}{c}{Ground Pen}  \\
\textbf{Method} & \textbf{Input} & \textbf{Vis} & \textbf{Occ} & \textbf{All} & \textbf{Legs} & \textbf{Vtx} & \textbf{Contact} & \textbf{Accel} & \textbf{Freq} & \textbf{Dist} \\
\hline
VPoser-t & Occ Keypoints & \textbf{0.67} & 20.76 & 9.22 & 21.08 & 7.95 & - & 5.71 & 16.77\% & 2.28 \\
MVAE~\cite{ling2020MVAE} & Occ Keypoints &  2.39 & 19.15 & 9.52 & 16.86 & 8.90 & - & 7.12 & \textbf{3.15\%} & 0.30 \\
\methodname (Ours) & Occ Keypoints & 1.46 & \textbf{17.40} & \textbf{8.24} & \textbf{15.42} & \textbf{7.56} & 0.89 & \textbf{5.38} & 3.31\% & \textbf{0.26} \\
\hline
VPoser-t & Noisy Joints & - & - & 3.67 & 4.47 & 4.98 & - & \textbf{4.61} & 1.35\% & 0.07 \\
MVAE~\cite{ling2020MVAE} & Noisy Joints & - & - & 2.68 & 3.21 & 4.42 & - & 6.5 & 1.75\% & 0.11 \\
\methodname (Ours) & Noisy Joints & - & - &  \textbf{2.27} & \textbf{2.61} & \textbf{3.55} & 0.97 & 5.23 & \textbf{1.18\%} & \textbf{0.05}
\end{tabular}
\end{center}
\caption{Motion and shape estimation from 3D observations: partially occluded keypoints (top) and noisy joints (bottom). \emph{Positional Error} ($cm$) is reported w.r.t. the input modality. Acceleration is $m/s^2$ and penetration distance in $cm$.\vspace{-5mm}}
\label{table:3dresults}
\end{table*}

\subsection{Generative Model Evaluation}
\label{expt:gen}
We first evaluate \methodname as a standalone generative model and show improved generalization to unseen motions and bodies compared to \emph{MVAE} for two common tasks (see Table~\ref{table:motionprior}): future prediction and diverse sampling. We use $2s$ AMASS sequences and start generation from the first step. Results are shown for \emph{\methodname} and a modified \emph{\methodname (Qual)} that uses $\smpljoints$ as input to each step during rollout instead of $\joints$, thereby enforcing skeleton consistency. This version produces \emph{qualitatively} superior results for generation, but is too expensive to use during \optname. 

For prediction, we report average displacement error (\textbf{ADE}) and final displacement error (\textbf{FDE})~\cite{yuan2020dlow}, which measure mean joint errors over all steps and at the final step, respectively. We sample 50 $2s$ motions for each initial state and the one with lowest \textbf{ADE} is considered the prediction. For diversity, we sample 50 $5s$ motions and compute the average pairwise distance (\textbf{APD})~\cite{aliakbarian2020stochastic}, \ie the mean joint distance between all pairs of samples.

As seen in~\cref{table:motionprior}, the base MVAE~\cite{ling2020MVAE} does not generalize well when trained on the large AMASS dataset; our proposed CVAE improves both the accuracy and diversity of samples. \emph{\methodname (Qual)} hinders prediction accuracy, but gives better diversity and visual quality (see supplement). %

\begin{figure}
\begin{center}
\includegraphics[width=1.0\columnwidth]{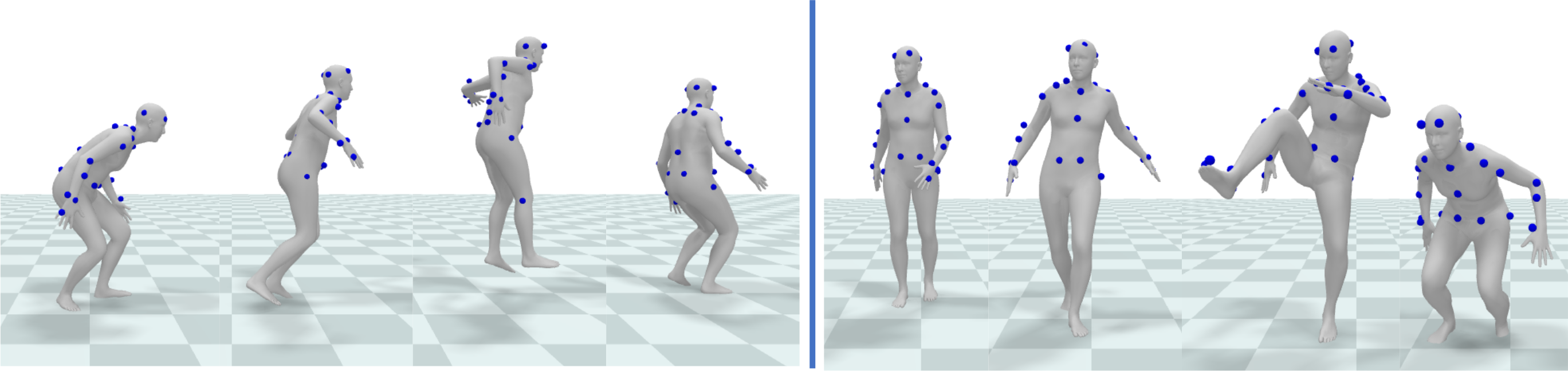}
\end{center}
\vspace{-2mm}
\captionof{figure}{Fitting to partial 3D \textcolor{blue}{keypoints}. \methodname captures non-periodic motions like jumping, crouching, and kicking.\vspace{-5mm}}
\label{fig:qual3dkey}
\end{figure}

\subsection{Estimation from 3D Observations}
\label{expt:estim3d}
Next, we show that \methodname also generalizes better when used in \optname for fitting to 3D data, and that using a motion prior is crucial to plausibly handling occlusions. $3s$ AMASS sequences are used to demonstrate key abilities: (i) fitting to partial data and (ii) denoising. For the former, \optname fits to 43 keypoints on the body that resemble motion capture markers; keypoints that fall below $0.9m$ at each timestep are ``occluded'', leaving the legs unobservable at most steps. For denoising, Gaussian noise with $4cm$ standard deviation is added to 3D joint position observations.

\cref{table:3dresults} compares to \emph{VPoser-t} and to using \emph{MVAE} as the motion prior during optimization rather than \methodname. We report leg joint errors (toes, ankles, and knees), which are often occluded, separately. The right side of the table reports plausibility metrics. \methodname gives more accurate poses, especially for occluded keypoints and leg joints. It also estimates smoother motions with fewer and less severe ground penetrations. For denoising, \emph{VPoser-t} oversmooths which gives the lowest acceleration but least accurate motion. \optname with \methodname gives inherently smooth results while still allowing for necessarily large accelerations to fit dynamic observations. Notably, \methodname predicts person-ground contact with 97\% accuracy even under severe noise. Qualitative results are shown in \cref{fig:teaser} and \cref{fig:qual3dkey}. %

\begin{figure}
\begin{center}
\includegraphics[width=1.0\columnwidth]{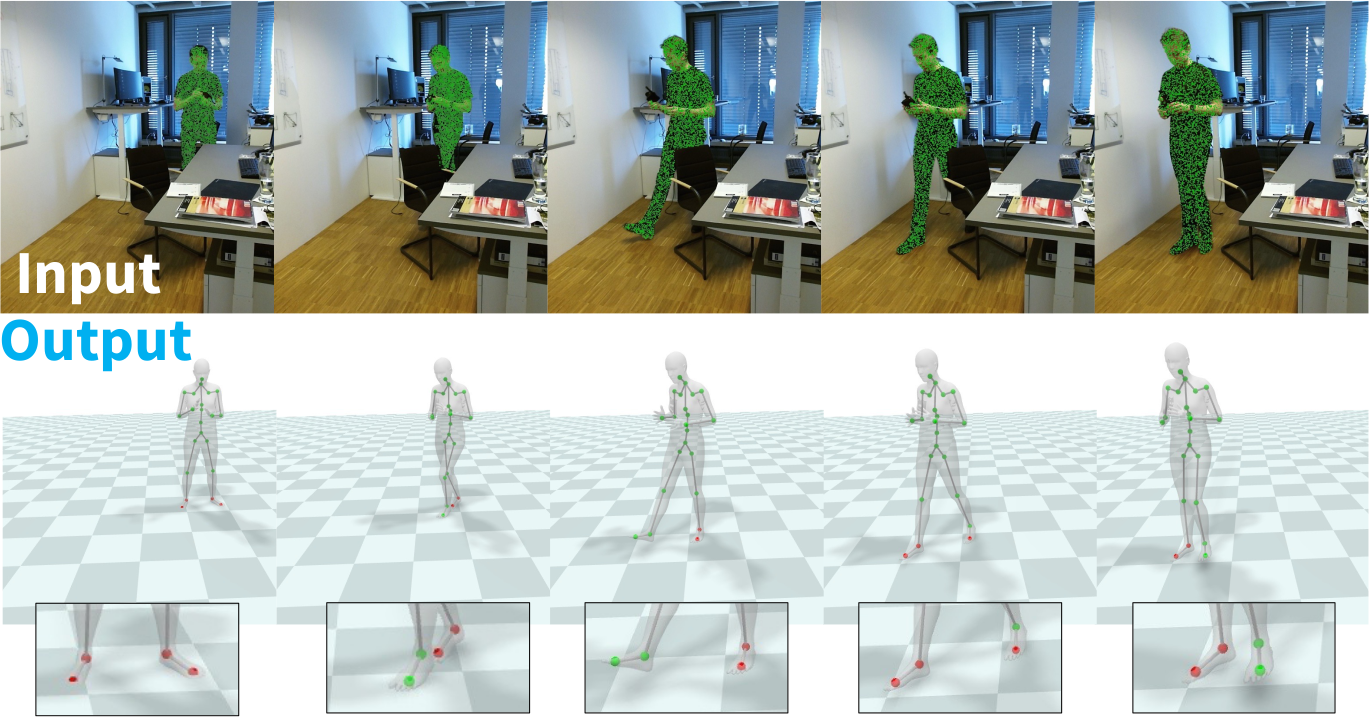}
\end{center}
\captionof{figure}{From RGB-D (top) \optname with \methodname outputs 3D motion, the ground plane, and \textcolor{red}{contacts} (bottom).\vspace{-3mm}}
\label{fig:qual3ddepth}
\end{figure}

\begin{table*}
\setlength{\tabcolsep}{7pt}
\begin{center}
\begin{tabular}{l | c c c c | c c c c | c c c}
 & \multicolumn{4}{c|}{Global Joint Error}  & \multicolumn{4}{c|}{Root-Aligned Joint Error}  & & \multicolumn{2}{c}{Ground Pen} \\
\textbf{Method} & \textbf{Vis} & \textbf{Occ} & \textbf{All} & \textbf{Legs} & \textbf{Vis} & \textbf{Occ} & \textbf{All} &  \textbf{Legs} & \textbf{Accel} & \textbf{Freq} & \textbf{Dist} \\
\hline
VIBE~\cite{kocabas2019vibe} & 90.05 & 192.55 & 116.46 & 121.61 & 12.06 & 23.78 & 15.08 & 21.65 & 243.36 & 7.98\% & 3.01 \\
VPoser-t  & 28.33 & 40.97 & 31.59 & 35.06 & 12.77 & 26.48 & 16.31 & 25.60 & 4.46 & 9.28\% & 2.42 \\
MVAE~\cite{ling2020MVAE}  & 37.54 & 50.63 & 40.91 & 44.42 & 16.00 & 28.32 & 19.17 & 26.63 & 4.96 & 7.43\% & 1.55 \\
\hline
No Delta & 27.55 & 35.59 & 29.62 & 32.14 & \textbf{11.92} & 23.10 & 14.80 & 21.65 & 3.05 & 2.84\% & \textbf{0.58} \\
No Contacts & 26.65 & 39.21 & 29.89 & 35.73 & 12.24 & 23.36 & 15.11 & 22.25 & \textbf{2.43} & 5.59\% & 1.70 \\
No $\loss{SMPL}$ & 31.09 & 43.67 & 34.33 & 36.84 & 12.81 & 25.47 & 16.07 & 23.54 & 3.21 & 4.12\% & 1.31 \\
Standard Prior & 77.60 & 146.76 & 95.42 & 99.01 & 18.67 & 39.40 & 24.01 & 34.02 & 5.98 & 8.30\% & 6.47 \\
\hline
\methodname (Ours) &  \textbf{26.00} & \textbf{34.36} & \textbf{28.15} & \textbf{31.26} & 12.02 & \textbf{21.70} & \textbf{14.51} & \textbf{20.74} & \textbf{2.43} & \textbf{2.12\%} & 0.68 
\end{tabular}
\end{center}
\caption{Motion and shape from RGB video (\ie 2D joints) on i3DB~\cite{monszpart2019iMapper}. Joint errors are in $cm$ and acceleration is $m/s^2$. Top shows results from motion estimation baselines while bottom uses ablations of \methodname during optimization.}
\label{table:imapper}
\end{table*}
\begin{figure*}[t]
\begin{center}
\includegraphics[width=1.0\textwidth]{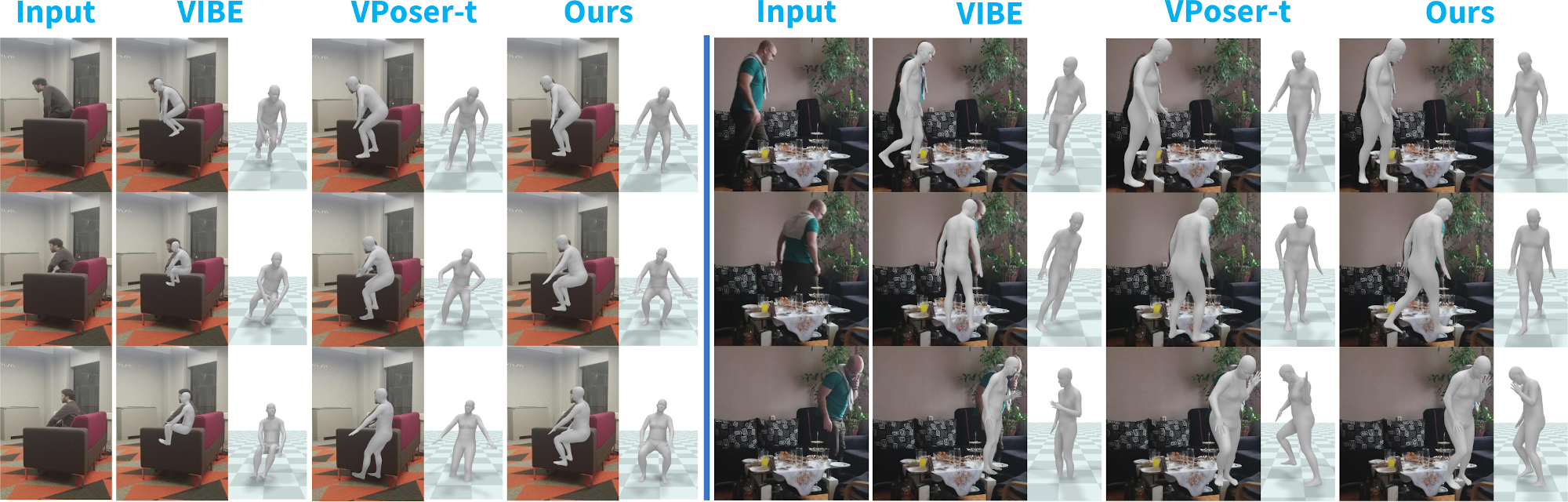}
\end{center}
\captionof{figure}{Qualitative comparison for fitting to RGB video (\ie 2D joints) from i3DB~\cite{monszpart2019iMapper}. Optimization using \methodname (Ours) outputs natural and plausible sitting and walking motions under heavy occlusions compared to baseline approaches.\vspace{-3mm}}
\label{fig:imapperres}
\end{figure*}

\begin{table}
\setlength{\tabcolsep}{6pt}
\begin{center}
\begin{tabular}{l | c | c c c}
 & & & \multicolumn{2}{c}{Ground Pen} \\
\textbf{Method} & \textbf{Input} & \textbf{Accel} & \textbf{Freq} & \textbf{Dist} \\
\hline
VIBE~\cite{kocabas2019vibe} & RGB & 86.06 & 23.46\% & 4.71 \\
PROX-RGB~\cite{hassan2019prox} & RGB &  196.07 &\textbf{ 2.55\%} & \textbf{0.32} \\
VPoser-t & RGB & 3.14 & 13.38\% & 2.82  \\
\methodname (Ours) & RGB & \textbf{1.73} & 9.99\% & 1.56 \\
\hline
PROX-D~\cite{hassan2019prox} & RGB-D & 46.59 & 8.95\% & 1.19 \\
VPoser-t & RGB-D & 3.27 & 10.66\% & 2.18 \\
\methodname (Ours) & RGB-D &\textbf{1.61} & \textbf{5.19\%} & \textbf{0.85}
\end{tabular}
\end{center}
\caption{Plausibility evaluation on videos in PROX~\cite{hassan2019prox}. Acceleration is $m/s^2$ and penetration distance in $cm$.\vspace{-3mm}}
\label{table:prox}
\end{table}

\subsection{Estimation from RGB(-D) Observations}
\label{expt:estim2d}
Finally, we show that \optname with \methodname can be applied to real-world RGB and RGB-D observations, and outperforms baselines on positional and plausibility metrics especially from partial and noisy data. We use $3s$ (90 frame) clips from i3DB~\cite{monszpart2019iMapper} and PROX~\cite{hassan2019prox}. \cref{table:imapper} shows results on i3DB which affords quantitative 3D joint evaluation. The top half compares to baseline estimation methods; the bottom uses ablations of \methodname in \optname rather than the full model. Mean per-joint position errors are reported for \textbf{global} joint positions and after root alignment.

As seen in \cref{table:imapper}, \emph{VIBE} gives locally accurate predictions for visible joints, but large global errors and unrealistic accelerations due to occlusions and temporal inconsistency (see~\cref{fig:imapperres}). \emph{VPoser-t} gives reasonable global errors, but suffers frequent penetrations as shown for sitting in~\cref{fig:imapperres}. Using \emph{MVAE} or ablations of \methodname as the motion prior in \optname fails to effectively generalize to real-world data and performs worse than the full model. The conditional prior and $\loss{SMPL}$ have the largest impact, while performance even without using contacts still outperforms the baselines.

The top half of~\cref{table:prox} evaluates plausibility on additional RGB results from PROX compared to \emph{VIBE} and \emph{PROX-RGB}. Since \emph{PROX-RGB} uses the scene mesh as input to enforce environment constraints, it is a very strong baseline and its performance on penetration metrics is expectedly good. \methodname comparatively increases penetration frequency since it only gets a rough ground plane as initialization, but gives much smoother motions.

The bottom half of~\cref{table:prox} shows results fitting to RGB-D for the same PROX data, which uses both $\ensub{data}{2D}$ \emph{and} $\ensub{data}{PC3D}$ in \optname. This improves performance using \methodname, slightly outperforming $\emph{PROX-D}$ which is less robust to issues with 2D joint detections and 3D point noise causing large errors. Qualitative examples are in \cref{fig:teaser} and \cref{fig:qual3ddepth}. %

Thanks to the generalizability of HuMoR, TestOpt is also effective in recovering very dynamic motions like dancing from RGB video when the full body is visible (see supplementary material for examples).
\section{Discussion}
We have introduced \methodname, a learned generative model of 3D human motion leveraged during test-time optimization to robustly recover pose and shape from 3D, RGB, and RGB-D observations. We have demonstrated that the key components of our model enable generalization to novel motions and body shapes for both generative tasks and downstream optimization. Compared to strong learning and optimization-based baselines, \methodname excels at estimating plausible motion under heavy occlusions, and simultaneously produces consistent ground plane and contact outputs. 

\paragraph{Limitations \& Future Work}
\methodname leaves ample room for future studies. The static camera and ground plane assumptions are reasonable for indoor scenes but true \emph{in-the-wild} operation demands methods handling dynamic cameras and complex terrain. Our rather simplistic contact model should be upgraded to capture scene-person interactions for improved motion and scene perception. Lastly, we plan to \emph{learn} motion estimation directly from partial observations which will be faster than TestOpt and enable \emph{sampling} multiple plausible motions rather than relying on a single local minimum.

{\small \paragraph{Acknowledgments} 
This work was supported by the Toyota Research Institute (``TRI") under the University 2.0 program, grants from the Samsung GRO program and the Ford-Stanford Alliance, a Vannevar Bush faculty fellowship, NSF grant IIS-1763268, and NSF grant CNS-2038897. TRI provided funds to assist the authors with their research but this article solely reflects the opinions and conclusions of its authors and not TRI or any other Toyota entity.}

{\small
\bibliographystyle{ieee_fullname}
\bibliography{refs,gpscpaper}
}

\clearpage
\appendix
\section*{Appendices}
Here we provide details and extended evaluations omitted from the main paper\footnote{In the rest of this document we refer to the main paper briefly as \emph{paper}.} for brevity. \cref{supp:sec:discussions} provides extended discussions, 
\cref{supp:sec:cvae} and \cref{supp:sec:testopt} give method details regarding the \methodname model and test-time optimization (\optname), 
\cref{supp:sec:map} derives our optimization energy from a probabilistic perspective, \cref{supp:sec:exptdetails} provides experimental details from the main paper, and \cref{supp:sec:evals} contains extended experimental evaluations.

We encourage the reader to view the \textbf{supplementary videos} on the project webpage\footnote{\href{https://geometry.stanford.edu/projects/humor/}{https://geometry.stanford.edu/projects/humor/}} and supplementary webpage\footnote{\href{https://geometry.stanford.edu/projects/humor/supp.html}{https://geometry.stanford.edu/projects/humor/supp.html}} for extensive qualitative results. We further discuss these results in \cref{supp:sec:evals}.

\section{Discussions}
\label{supp:sec:discussions}
\paragraph{State Representation}
Our state representation is somewhat redundant to include both explicit joint positions $\joints$ and SMPL parameters (which also give joint positions $\smpljoints$). This is motivated by recent works~\cite{ling2020MVAE,zhang2020mojo} which show that using an extrinsic representation of body keypoints (\eg joint positions or mesh vertices) helps  in learning motion characteristics like static contact, thereby improving the visual quality of generated motions. The over-parameterization, unique to our approach, additionally allows for consistency losses leveraged during CVAE training and in \optname.

Another noteworthy property of our state is that it does not explicitly represent full-body shape -- only bone proportions are implicitly encoded through joint locations. During training, we use shape parameters $\shape$ provided in AMASS~\cite{mahmood2019amass} to compute $\loss{SMPL}$, but otherwise the CVAE is shape-unaware. Extending our formulation to include full-body shape is an important direction for improved generalization and should be considered in future work.

\paragraph{Conditioning on More Time Steps}
Alternatively, we could condition the dynamics learned by the CVAE with additional previous steps, \ie $\ptheta(\xt | \xtm , \dots, \mathbf{x}_{t-p})$, however since $\xtm$ includes velocities this is unnecessary and only increases the chances of \emph{overfitting} to training motions. It would additionally increases the necessary computation for both generation and \optname. %

\paragraphnoper{Why CVAE?}
Our use of a CVAE to model motion is primarily motivated by recent promising results in the graphics community~\cite{ling2020MVAE,ghorbani2020probabilistic}. Not only is it a simple solution, but also affords the physical interpretation presented in the main paper. Other deep generative models could be considered for $\ptheta(\xt | \xtm)$, however each have potential issues compared to our CVAE. The conditional generative adversarial network~\cite{mirza2014conditional} would use standard normal noise for $\zt$, which we show is insufficient in multiple experiments. Furthermore, it does not allow for inferring a latent transition $\zt$. Past works have had success with recurrent and variational-recurrent architectures~\cite{zhang2020mojo}. As discussed previously, the reliance of these networks on multiple timesteps increases overfitting which is especially dangerous for our estimation application which requires being able to represent arbitrary observed motions. Finally, normalizing flows~\cite{KobyzevPAMI2020} and neural ODEs~\cite{chen2018neural} show exciting potential for modeling human motion, however conditional generation with these models is not yet well-developed.

\paragraph{In comparison to Motion VAE (MVAE)~\cite{ling2020MVAE}}
Our proposed CVAE is inspired by MVAE, but introduces a number of key improvements that enable generalization and expressivity: (\textbf{i}) \methodname uses a neural network to learn a conditional prior $\prior$ rather than assuming $\ptheta(\zt) = \normal (\zt; \mathbf{0}, \mathbf{I})$, (\textbf{ii}) the decoder predicts the \emph{change in state} $\Delta_\theta$ rather than the next state $\xt$ directly, (\textbf{iii}) the decoder outputs person-ground contacts $\mathbf{c}_t$ which MVAE does not model, and (\textbf{iv}) \methodname trains using $\loss{SMPL}$ regularization to encourage joint position/angle consistency whereas MVAE uses the typical ELBO. Tab.~3 in the main paper shows that these differences are crucial to achieve good results as MVAE does not work well. The conditional prior and $\loss{SMPL}$ are particularly important: learning $\prior$ is theoretically justified when deriving the CVAE and is intuitive for human motion, while $\loss{SMPL}$ provides a strong supervision to improve stability of model rollout. 

Furthermore, the pose state employed by \methodname and MVAE differ slightly. In MVAE, the root is defined by projecting the pelvis onto the ground, giving 2D linear and 1D angular velocities. In HuMoR, the root is at the pelvis giving full 3D velocities.

\paragraph{A Note on $\beta$-VAE~\cite{higgins2016beta}}
The KL weight $\w{KL}$ in Eq. 7 of the main paper is not directly comparable to a typical $\beta$-VAE~\cite{higgins2016beta} due to various implementation details. First, $\loss{rec}$ is the mean-squared error (MSE) of the unnormalized state rather than the true log-likelihood. The use of additional regularizers $\loss{reg}$ that are not formulated probabilistically to be part of the reconstruction loss further compounds the difference. Furthermore, in practice losses are averaged over both the feature and batch dimensions as not to depend on chosen dimensionalities. All these differences result in setting $\w{KL} = 4\e{-4}$. 

\paragraph{The Need for $\en{reg}$ in Optimization}
The motion prior term $\en{mot}=\en{CVAE}+\en{init}$, which leverages our learned conditional prior and GMM, nicely falls out of the MAP derivation (see \cref{supp:sec:map} below) and is by itself reasonable to ensure motion is plausible. However, in practice it can be prone to local minima and slow to converge without any regularization. This is primarily because \methodname is trained on clean motion capture data from AMASS~\cite{mahmood2019amass}, but in the early stages of optimization the initial state $\xzero$ will be far from this domain. This means rolled out motions using the CVAE decoder will be implausible and the likelihood output from learned conditional prior is not necessarily meaningful (since inputs will be well outside the training distribution). The additional regularizers presented in the main paper, mainly $\en{skel}$ and $\en{env}$, allow us to resolve this issue by reflecting expected behavior of the motion model when it is producing truly plausible motions (\ie $\xzero$ is similar to the training data).

\paragraph{On Evaluation Metrics}
As discussed in prior work~\cite{RempeContactDynamics2020}, traditional positional metrics used to evaluate root-relative pose estimates do not capture the accuracy of the absolute (``global'') motion nor its physical/perceptual plausibility. This is why we use a range of metrics to capture both the global joint accuracy, local joint accuracy (after aligning root joints), and plausibility of a motion. However, these metrics still have flaws and there is a need to develop more informative motion estimation evaluation metrics for both absolute accuracy and plausibility. This is especially true in scenarios of severe occlusions where there is not a single correct answer: even if the ``ground truth'' 3D joints are available, there may be multiple motions that explain the partial observations equally well.

\paragraph{On Convergence} Our multi-objective optimization uses a mixture of convex and non-convex loss functions. As we utilize L-BFGS, the minimum energy solution we report is only locally optimal. While simulated annealing or MCMC / HMC (Markov Chain Monte Carlo / Hamiltonian Monte Carlo) type of exploration approaches can be deployed to search for the global optimum, such methods would incur heavy computational load and hence are prohibitive in our setting. Thanks to the accurate initialization, we found that most of the time \optname converges to a good minimum.
This observation is also lightly supported by recent work arguing that statistically provable convergence can be attained for the human pose problem under convex and non-convex regularization using a multi-stage optimization scheme~\cite{wangni2021towards}. 

\subsection{Assumptions and Limitations}
\label{supp:sec:failurecases}
\paragraph{On the Assumption of a Ground Plane}
We use \emph{the ground} during \optname to obtain a transformation to the canonical reference frame where our prior is trained. While this is a resonable assumption in a majority of scenarios, we acknowledge that certain applications might require \emph{in-the-wild} operation where a single ground plane does not exist \eg climbing up stairs or moving over complex terrain. 
In such scenarios, we require a consistent reference frame, which can be computed from: (i) an accelerometer if a mobile device is used, (ii) pose of static, rigid objects if an object detector is deployed, (iii) fiducial tags or any other means of obtaining a gravity direction.

Note that the ground plane is not an essential piece in the test-time optimization. It is a requirement only because of the way our CVAE is trained: on motions with a ground plane at $z=0$, gravity in the $-z$ direction, and without complex terrain interactions.
Although we empirically noticed that convergence of training necessitates this assumption, other architectures or the availability of larger \emph{in-the-wild} motion datasets might make training \methodname possible under arbitrary poses. This perspective should clarify why our method \emph{can work} when the ground is invisible: \optname might converge from a bad initialization as long as our prior (\methodname) is able to account for the observation. 

\paragraph{On the Assumption of a Static Camera}
While a static camera is assumed in all of our evaluations, recent advances in 3D computer vision make it possible to overcome this limitation. Our method, backed by either a structure from motion / SLAM pipeline or a camera relocalization engine, can indeed work in scenarios where the camera moves as well as the human targets. A more sophisticated solution could leverage our learned motion model to disambiguate between camera and human motion. Expectedly, this requires further investigation, making room for future studies as discussed at the end of the main paper.

\begin{figure*}
\begin{center}
\includegraphics[width=\linewidth]{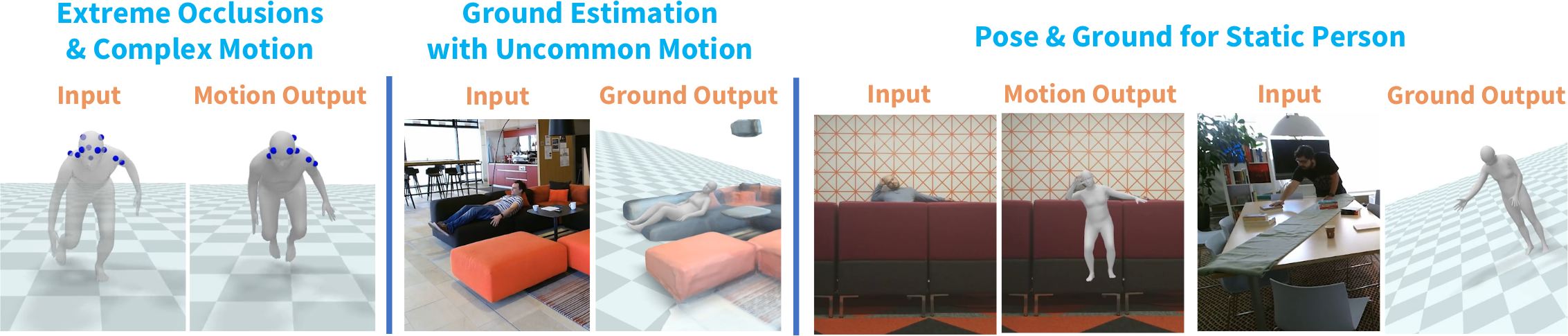}
\end{center}
\captionof{figure}{Failure cases of \optname using \methodname. Please see \cref{supp:sec:failurecases} or the supplementary videos for details of each.} %
\label{fig:qual-failures}
\end{figure*}

\paragraph{Other Limitations and Failure Cases}
As discussed in Sec. 6 of the main paper, \methodname has limitations that motivate multiple future directions. 
First, optimization is generally slow compared to learning-based (direct prediction) methods. This also reflects on our test-time optimization. Approaches for \emph{learning to optimize} can come handy in increasing the efficiency of our method.
Additionally, our current formulation of \optname allows only for a single output, the local optimum. Therefore, future work may explore learned approaches yielding multi-hypothesis output, which can be used to characterize uncertainty.

Specific failure cases (as shown in the supplementary videos and \cref{fig:qual-failures}) further highlight areas of future improvement. First, extreme occlusions (e.g. only a few visible points as in \cref{fig:qual-failures} left), especially at the first frame which determines $\x_0$, makes for a difficult optimization that often lands in local minima with implausible motions. Second, uncommon motions that are rare during CVAE training, such as laying down in \cref{fig:qual-failures} (middle), can cause spurious ground plane outputs as \optname attempts to make the motion more likely. Leveraging more holistic scene understanding methods and models of human-environment interaction will help in these cases. Finally, our method is dependent on motion in order to resolve ambiguity, which is usually very helpful but has corner cases as shown in \cref{fig:qual-failures} (right). For example, if the observed person is nearly static, the optimization may produce implausible poses due to ambiguous occlusions (\eg standing when really the person is sitting) and/or incorrect ground plane estimations.

\section{\methodname Model Details}
\label{supp:sec:cvae}
In this section, we provide additional implementation details for the \methodname motion model described in Sec. 3 of the main paper.
\subsection{CVAE Architecture and Implementation}
\paragraph{Body Model}
We use the SMPL+H body model~\cite{smplh:SIGGRAPHASIA:2017} since it is used by the AMASS~\cite{mahmood2019amass} dataset. However, our focus is on modeling body motion, so \methodname and \optname do not consider the hand joints (leaving the 22 body joints including the root). Hand joints could be straightforwardly optimized with body motion, but was not in our current scope.

\paragraph{Canonical Coordinate Frame} To ease learning and improve generalization, our network operates on inputs in a canonical coordinate frame. Specifically, based on $\xtm$ we apply a rotation around the up ($+z$) axis and translation in $x,y$ such that the $x$ and $y$ components of $\trans_{t-1}$ are 0 and the person's body right axis (w.r.t. $\orient_{t-1}$) is facing the $+x$ direction.

\paragraph{Architecture}
The encoder and prior networks are identical multi-layer perceptrons (MLP) with 5 layers and hidden size 1024. The decoder is a 4-layer MLP with hidden sizes (1024, 1024, 512). The latent transition $\zt \in \reals^{48}$ is skip-connected to every layer of the decoder in order to emphasize its importance and help avoid posterior collapse~\cite{ling2020MVAE}.  ReLU non-linearities and group normalization~\cite{wu2018group} with 16 groups are used between all layers except outputs in each network. Input rotations are represented as matrices, while the network outputs the axis-angle representation in $\RThree$. In total, the CVAE network contains $\sim$9.7 million parameters.

\subsection{CVAE Training}
\label{supp:sec:cvaetraining}
\paragraph{Losses}
The loss function used for training is primarily described in the main paper (see Eq. 7). For a training pair $(\xtm, \xt)$, the KL divergence loss term is computed between the output distributions of the encoder and conditional prior as 
\begin{align}
    \loss{KL} &= \infdiv{\encoder}{\prior} \nonumber \\
              &= D_\text{KL}( \normal (\zt; \mu_\phi (\xt, \xtm), \sigma_\phi (\xt, \xtm))  \nonumber \\ 
               & \qquad\quad ||  \normal (\zt; \mu_\theta(\xtm), \sigma_\theta(\xtm)) ).
\end{align}

The SMPL loss $\loss{SMPL}$ is computed using the ground truth shape parameters $\shape$ provided in AMASS on the ground truth gendered body model. 

\paragraph{Dataset} 
For training, we use AMASS~\cite{mahmood2019amass}: a large, publicly-available motion capture (mocap) database containing over 11k motion sequences from 344 different people fit to SMPL. The database aggregates and standardizes many mocap datasets into one. We pre-process AMASS by cropping the middle 80\% of each motion sequence, sub-sampling to 30 Hz, estimating velocities with finite differences, and using automated heuristics based on foot contacts to remove sequences with substantial terrain interaction (\eg stairs, ramps, or platforms). We automatically annotate ground contacts for 8 body joints (\emph{left and right toes}, \emph{heels}, \emph{knees}, and \emph{hands}) based on velocity and height. In particular, if a joint has moved less than $0.5cm$ in the last timestep and its $z$ component is within $8cm$ of the floor, it is considered to be in contact. For toe joints, we use a tighter height threshold of $4cm$.

For training the CVAE, we use the recommended training split (save for TCD Hands~\cite{TCD_hands} which contains mostly hand motions): CMU~\cite{cmuWEB}, MPI Limits~\cite{PosePrior_Akhter:CVPR:2015}, TotalCapture~\cite{TotalCaptureTrumble:BMVC:2017}, Eyes Japan~\cite{Eyes_Japan}, KIT~\cite{KIT_Dataset}, BMLrub~\cite{BMLrub}, BMLmovi~\cite{ghorbani2020movi}, EKUT~\cite{KIT_Dataset}, and ACCAD~\cite{ACCAD}. 
For validation during training we use MPI HDM05~\cite{MPI_HDM05}, SFU~\cite{SFU}, and MPI MoSh~\cite{MoSh_lopermahmoodetal2014}. Finally for evaluations (Sec. 5.3 of the main paper), we use HumanEva~\cite{sigal2010humaneva} and Transitions~\cite{mahmood2019amass}.

\paragraph{Training Procedure}
We train using 10-step sequences sampled on-the-fly from the training set (in order to use scheduled sampling as detailed below). To acquire a training sequence, a full mocap sequence is randomly (uniformly) chosen from AMASS and then a random 10-step window within that sequence is (uniformly) sampled. Training is performed using batches of 2000 sequences for 200 epochs with Adamax~\cite{kingma2014adam} and settings $\beta_1 = 0.9$, $\beta_2=0.999$, and $\epsilon=1e^{-8}$. We found this to be more stable than using Adam. The learning rate starts at $1e^{-4}$ and decays to $5e^{-5}$, $2.5e^{-5}$, and $1.25e^{-5}$ at epochs 50, 80, and 140, respectively. We use early stopping by choosing the network parameters that result in the best validation split performance throughout training. 

A common difficulty in training VAEs is posterior collapse~\cite{lucas2019collapse} -- when the learned latent encoding $\zt$ is effectively ignored by the decoder. This problem is exacerbated in CVAEs since the decoder receives additional conditioning~\cite{ling2020MVAE,sohn2015CVAE}. To combat collapse, we linearly anneal $\w{KL}$ from $0.0$ to its full value of $4e^{-4}$ over the first 50 epochs. We also found that our full model, which uses a learned conditional prior, was less susceptible to posterior collapse than the baselines that assume $\ptheta(\zt) = \normal (\zt; \mathbf{0}, \mathbf{I})$.

\paragraph{Training Computational Requirements}
We train our CVAE on a single Tesla V100 16GB GPU, which takes approximately 4 days.

\paragraph{Scheduled Sampling}
As explained in the main paper, our scheduled sampling follows~\cite{ling2020MVAE}.
In particular, at each training epoch $i$ we define a probability $\schedsamp \in [0.0, 1.0]$ of using the ground truth state input $\xtm$ at each timestep $t$ in a training sequence, as opposed to the model's own previous output $\pxtm$. Training is done using a curriculum that includes $\schedsamp = 1.0$ (regular supervised training), $\schedsamp \in (0.0, 1.0)$ (mix of true and self inputs at each step), and finally $\schedsamp = 0.0$ (always use full generated rollouts). Importantly for training stability, if using the model's own prediction $\pxtm$ as input to $t$, we do not backpropagate gradients from the loss on $\pxt$ back through $\pxtm$. 

For CVAE training, we use 10 epochs of regular supervised training, 10 of mixed true and self inputs, and the rest using full self-rollouts.

\subsection{Initial State GMM}
\paragraph{State Representation}
Since the GMM models a single state, we use a modified representation that is minimal (\ie avoids redundancies) in order to be useful during test-time optimization. In particular the GMM state is
\begin{align}
\label{eqn:gmmstate}
    \x^\text{GMM} = [ \quad \transvel \quad \orientvel \quad \joints \quad \jointsvel \quad ]
\end{align}
with $\transvel, \orientvel \in \RThree$ the root linear and angular velocities, and joint positions and velocities $\joints,\jointsvel \in\reals^{3\times 22}$. During \optname, joints are determined from the current SMPL parameters $\joints = \smpljoints = M(\trans,\orient,\pose,\shape)$ so that gradients of the GMM log-likelihood (Eq. 11 in the main paper) will be used to update the initial state SMPL parameters.

\paragraph{Implementation Details}
The GMM uses full covariance matrices for each of the 12 components and operates in the same canonical coordinate frame as the CVAE. It trains using expectation maximization\footnote{\href{https://scikit-learn.org/stable/modules/generated/sklearn.mixture.GaussianMixture.html}{using scikit-learn}} on every state in the same AMASS training set used for the CVAE.

\section{Test-Time Optimization Details}
\label{supp:sec:testopt}
In this section, we give additional details of the motion and shape optimization detailed in Sec. 4 of the main paper.

\paragraph{State Representation}
In practice, for optimizing $\xzero$ we slightly modify the state from Eq. 1 in the main paper. First, we remove the joint positions $\joints$ to avoid the previously discussed redundancy, which is good for training the CVAE but bad for test-time optimization. Instead, we use $\smpljoints$ whenever needed at $t_0$. Second, we represent body pose $\pose$ in the latent space of the VPoser~\cite{pavlakos2019smplx} pose prior with $\zpose_0 \in \mathbb{R}^{32}$. Whenever needed, we can map between full joint angles and latent pose using the VPoser encoder and decoder. Finally, in our implementation state variables $\xt$ are by default represented in the coordinate frame of the given observations, \eg relative to the camera, to allow easily fitting to data - they are transformed back into the canonical CVAE frame when necessary as discussed below.

\paragraph{Floor Parameterization}
As detailed in the main paper, to obtain the transformation between the canonical coordinate frame in which our CVAE is trained and the observation frame used for optimization, we additionally optimize the floor plane of the scene $\floor \in \RThree$. This parameterization is $\floor = d\hat{\mathbf{n}}$ where $\hat{\mathbf{n}}$ is the ground unit normal vector and $d$ the plane offset. To disambiguate the normal vector direction $\hat{\mathbf{n}}$ given $\floor$, we assume that the $y$-component of the normal vector must be negative, \ie it points upward in the camera coordinate frame. This assumes the camera is not severely tilted such that the observed scene is ``upside down''. 

\paragraph{Observation-to-Canonical Transformation}
We assume that gravity is orthogonal to the ground plane. Therefore, given the current floor $\floor$ and root state $\trans,\orient$ (in the observation frame) we compute a rotation and translation to the canonical CVAE frame: after the transformation, $\hat{\mathbf{n}}$ is aligned with $+z$ and $d = 0$, $\orient$ faces body right towards $+x$, and the $x,y$ components of $\trans$ are 0. With this ability, we can always compute the (observed) state at time $\xt$ from $\z_{1:t}$, $\xzero$, and $\floor$ by (i) transforming $\xzero$ to the canonical frame, (ii) using the CVAE to rollout $\xt=f(\xzero, \z_{1:t})$, and (iii) transforming $\xt$ back to the observation frame.

\paragraph{Optimization Objective Details}
The optimization objective is detailed in Sec. 4.2 of the main paper. To compensate for the heavy tailed behavior of real data, we use robust losses for multiple data terms. $\ensub{data}{2D}$ uses the Geman-McClure function~\cite{geman1987gmef} which for our purposes is defined as $\rho(r, \sigma) = (\sigma^2 r^2) / (\sigma^2 + r^2)$ for a residual $r$ and scaling factor $\sigma$. We use $\sigma=100$ for all experiments. $\ensub{data}{PC3D}$ uses robust bisquare weights~\cite{beaton1974fitting}. These weights are computed based on the one-way chamfer distance term (see Eq. 14 in the main paper): residuals over the whole sequence are first normalized using a robust estimate of the standard deviation based on the median absolute deviation (MAD), then each weight is computed as 
\begin{equation}
    w_\text{bs} = \begin{cases}
    (1 - (\hat{r} / \kappa)^2 )^2  & | \hat{r}/\kappa | < 1 \\
    0.0 & \text{else}
    \end{cases}
\end{equation}
In this equation, $\hat{r}$ is a normalized residual and $\kappa$ is a tuning constant which we set to 4.6851.

In the $\en{env}$ energy term, we use $\delta=8$ $cm$ to ensure the $z$-component of contacting joints are within $8$ $cm$ of the floor when in contact (since joints are inside the body) in the canonical frame.

\paragraph{Initialization}
As detailed in Sec. 4.2 of the main paper, our optimization is initialized by directly optimizing SMPL pose and shape parameters using $\en{data}$ and $\en{shape}$ along with a pose prior $\en{pose}$ and joint smoothing $\en{smooth}$. The latter are weighted by $\enw{pose}$ and $\enw{smooth}$. This two-stage initialization first optimizes global translation and orientation for 30 optimization steps, followed by full pose and shape for 80 steps. At termination, we estimate velocities using finite differences, which allows direct initialization of the state $\xzero^\text{init}$. To get $\zall^\text{init}$, the CVAE encoder is used to infer the latent transition between every pair of frames. The initial shape parameters $\shape_\text{init}$ are a direct output of the initialization optimization. Finally, for fitting to RGB(-D) the ground plane is initialized from video with PlaneRCNN~\cite{liu2019planercnn}, though we found simply setting the floor to $y=0$ (\ie the normal is aligned with the camera up axis) works just as well in most cases.

\paragraph{Optimization (\optname) Details}
Our optimization is implemented in PyTorch~\cite{paszke2017automatic} using L-BFGS with a step size of 1.0 and \emph{autograd}. For all experiments, we optimize using the neutral SMPL+H~\cite{smplh:SIGGRAPHASIA:2017} body model in 3 stages. First, only the initial state $\xzero$ and first 15 frames of the latent sequence $\z_{1:15}$ are optimized for 30 iterations in order to quickly reach a reasonable initial state. Next, $\xzero$ is fixed while the full latent dynamics sequence $\zall$ is optimized for 25 iterations, and then finally the full sequence and initial state are tuned together for another 15 iterations. The ground $\floor$ and shape $\shape$ are optimized in every stage.

The energy weights used for each experiment in the main paper are detailed in~\cref{table:energysettings}. The left part of the table indicates weights for the initialization phase (\ie the \emph{VPoser-t} baseline), while the right part is our full proposed optimization. A dash indicates the energy is not relevant for that data modality and therefore not used. Weights were manually tuned using the presented evaluation metrics and qualitative assessment. Note that for similar modalities (\eg 3D joints and keypoints, or RGB and RGB-D) weights are quite similar and so only slight tuning should be necessary to transfer to new data. The main tradeoff comes between reconstruction accuracy and motion plausibility: \eg the motion prior is weighted higher for i3DB, which contains many severe occlusions, than for PROX RGB where the person is often nearly fully visible.

\begin{table*}
\begin{center}
\setlength{\tabcolsep}{5pt}
\resizebox{\textwidth}{!}{
\begin{tabular}{l | c c c c c c | c c c c c c c c c c c}
& \multicolumn{6}{c|}{\textbf{Initialization}} & \multicolumn{11}{c}{\textbf{Full Optimization}} \\
\textbf{Dataset} & $\enw{data}^\text{3D}$ & $\enw{data}^\text{2D}$ & $\enw{data}^\text{PC3D}$ & $\enw{shape}$ & $\enw{pose}$ & $\enw{smooth}$ &  $\enw{data}^\text{3D}$ & $\enw{data}^\text{2D}$ & $\enw{data}^\text{PC3D}$ & $\enw{shape}$ & $\enw{CVAE}$ & $\enw{init}$ & $\enw{c}$ & $\enw{b}$ & $\enw{cv}$ & $\enw{ch}$ & $\enw{gnd}$ \\
\hline
AMASS (occ keypoints) &  1.0 & - & - & 0.015 & $2e^{-4}$ & 0.1 & 1.0 & - & - & 0.015 & $5e^{-4}$ & $5e^{-4}$ & 1.0 & 10.0 & 1.0 & 1.0 & - \\
AMASS (noisy joints) &  1.0 & - & - & 0.015 & $2e^{-4}$ & 10.0 & 1.0 & - & - & 0.015 & $1e^{-3}$ & $1e^{-3}$ & 1.0 & 10.0 & 1.0 & 1.0 & -  \\
i3DB (RGB) & - & $1e^{-3}$ & - & 4.5 & 0.04 & 100.0 & - & $1e^{-3}$ & - & 4.5 & 0.075 & 0.075 & 100.0 & $2e^{3}$ & 0.0 & 10.0 & 15.0 \\
PROX (RGB) & - & $1e^{-3}$ & - & 4.5 & 0.04 & 100.0 & - & $1e^{-3}$ & - & 4.5 & 0.05 & 0.05 & 100.0 & $2e^{3}$ & 100.0 & 10.0 & 15.0 \\
PROX (RGB-D) &  - & $1e^{-3}$ & 1.0 & 3.0 & 0.1 & 100.0 & - & $1e^{-3}$ & 1.0 & 3.0 & 0.075 & 0.075 & 100.0 & $2e^{3}$ & 100.0 & 10.0 & 90.0 \\
\end{tabular}
}
\end{center}
\caption{Energy weightings used in test-time optimization for each experiment in Sec. 5 of the main paper.\vspace{-3mm}}
\label{table:energysettings}
\end{table*}

\section{MAP Objective Derivation}
\label{supp:sec:map}
In this section, we formulate the core of the pose and shape optimization objective (Eq. 10 in the main paper) from a probabilistic perspective. Recall, we want to optimize the initial state $\xzero$, a sequence of latent variables $\zall$, ground $\floor$, and shape $\shape$ based on a sequence of observations $\obsall$. We are interested in the maximum a-posteriori (MAP) estimate:
\begin{align}
    &\max_{\xzero,\zall,\floor,\shape} p ( \xzero,\zall,\floor,\shape | \obsall ) \\
\label{eqn:map}
    = &\max_{\xzero,\zall,\floor,\shape} p ( \obsall | \xzero,\zall,\floor,\shape ) p (\xzero,\zall,\floor,\shape)
\end{align}
Assuming $\obszero$ is independent of $\floor$, the left term is written
\begin{align}
    p ( \obszero | \xzero,\shape ) \prod_{t=1}^{T} p( \obst | \zleq,\xzero,\floor,\shape)  
    = \prod_{t=0}^{T} p (\obst | \xt, \shape)\label{eqn:dataprob}
\end{align}
where $\mathbf{y}_t$ is assumed to only be dependent on the initial state and \textit{past} transitions. Additionally, $\{\zleq,\xzero,\floor\}$ is replaced with $\xt=f(\xzero, \z_{1:t})$ using CVAE rollout as detailed previously. 
The right term in~\cref{eqn:map} is written as 
\begin{align}
    p (\xzero,\floor,\shape) \prod_{t=1}^T p(\zt | \zprev, \xzero,\floor,\shape) \\
\label{eqn:priors}
    = p (\xzero | \floor) p(\floor) p(\shape) \prod_{t=1}^T p(\zt | \xtm)
\end{align}
where $\xzero$, $\zt$, and $\floor$ are assumed to be independent of $\shape$. We then use these results within~\cref{eqn:map} to optimize the log-likelihood:
\begin{flalign}
& \max_{\xzero,\zall,\floor,\shape} \, \log p ( \obsall | \xzero,\zall,\floor,\shape ) + \log p (\xzero,\zall,\floor,\shape) \nonumber \\
 = & \min_{\xzero,\zall,\floor,\shape} \,- \sum_{t=0}^{T} \log p (\obst | \xt, \shape) - \sum_{t=1}^T \log p(\zt | \xtm) \nonumber \\
 &    \,\,\qquad\quad- \log p (\xzero | \floor) - \log p(\floor) - \log p(\shape) \nonumber \\
  = & \min_{\xzero,\zall,\floor,\shape} \, \en{data} + \en{CVAE} + \en{init} + \en{gnd} + \en{shape} \\
= & \min_{\xzero,\zall,\floor,\shape} \, \en{mot} + \en{data} + \en{gnd} + \en{shape}.
\end{flalign}
Assuming each energy presented in the main paper can be written as the log-likelihood of a distribution, this formulation recovers our optimization objective besides the additional regularizers $\en{skel}$ and $\en{env}$ (these terms could, in principle, be written as part of a more complex motion prior term $\en{mot}$, however for simplicity we do not do this). Next, we connect each energy term as presented in Sec. 4.2 of the paper to the probabilistic perspective.

\paragraph{Motion Prior $\en{mot}$}
This term is already the log-likelihood of our \methodname motion model (Eq. 11 of the paper), which exactly aligns with the MAP derivation.

\paragraph{Data Term $\en{data}$}
The form of $p (\obst | \xt, \shape)$ is modality-dependent. In the simplest case the observations  $\obst$ are 3D joints (or keypoints with known correspondences) and $p (\obst | \xt, \shape)$ is defined by $\obst = \smpljoints_t + \epsilon$ with $\epsilon \sim \normal(0, \sigma_\text{data})$. Then the energy is as written in Eq. 12 of the paper. For other modalities (Eq. 13 and 14 in the paper), the data term can be seen as resulting from a more sophisticated noise model.

\paragraph{Ground Prior $\en{gnd}$}
We assume the ground should stay close to initialization so $p(\floor) = \normal(\floor; \floor^\text{init}, \sigma_\text{gnd})$ corresponding to the objective in the paper $\en{gnd} = \enw{gnd}|| \floor - \floor^\text{init} ||^2$.

\paragraph{Shape Prior $\en{shape}$}
The shape $\shape$ should stay near neutral zero and so $p(\shape) = \normal(\mathbf{0}, \mathbf{I})$ which gives the energy $\en{shape} = \enw{shape}|| \shape ||^2$.

\vspace{-1mm}\section{Experimental Evaluation Details}
\label{supp:sec:exptdetails}
In this section, we provide details of the experimental evaluations in Sec. 5 of the main paper.

\subsection{Datasets}
\paragraphnoper{AMASS~\cite{mahmood2019amass}}
We use the same processed AMASS dataset as described in~\cref{supp:sec:cvaetraining} for experiments. Experiments in Sec. 5.3 and 5.4 of the main paper use the held out Transitions and HumanEva~\cite{sigal2010humaneva} subsets which together contain 4 subjects and about 19 minutes of motion.

\paragraphnoper{i3DB~\cite{monszpart2019iMapper}} is a dataset of RGB videos captured at 30 Hz containing numerous person-environment interactions involving medium to heavy occlusions. It contains annotated 3D joint positions at 10 Hz along with a primitive cuboid 3D scene reconstruction. We run off-the-shelf 2D pose estimation (OpenPose) ~\cite{cao2019openpose}, person segmentation~\cite{chen2017rethinking}, and plane detection~\cite{liu2019planercnn} models to obtain inputs and initialization for our test-time optimization. We evaluate our method in Sec. 5.5 of the main paper on 6 scenes (scenes 5, 7, 10, 11, 13, and 14) containing 2 people which totals about 1800 evaluation frames. From the annotated 3D objects, we fit a ground plane which is used to compute plausibility metrics.

\paragraphnoper{PROX~\cite{hassan2019prox}} is a large-scale dataset of RGB-D videos captured at 30 Hz containing person-scene interactions in a variety of environments with light to medium occlusions. We use a subset of the qualitative part of the dataset to evaluate the plausibility of our method's estimations. The data does \emph{not} have pose annotations, but does contain the scanned scene mesh to which we fit a ground plane for plausibility evaluation. We obtain 2D pose, person masks, and ground plane initialization in the same way as for i3DB. We evaluate in Sec. 5.5 of the main paper on all videos from 4 chosen scenes (N3Office, N3Library, N0Sofa, and MPH1Library) that tend to have more dynamic motions and occlusions. In total, these scenes contain 12 unique people and about 19 minutes of video.

\subsection{Baselines and Evaluation Metrics}

\paragraph{Motion Prior Baselines}
To be usable in our whole framework (\eg test-time optimization with SMPL), the \emph{MVAE} baseline is our proposed CVAE with all ablations applied simultaneously (no delta step prediction, no contact prediction, no SMPL losses, and no learned conditional prior). Note that this differs slightly from the model as presented in~\cite{ling2020MVAE}: the decoder is an MLP rather than a mixture-of-experts and the layer sizes are larger to provide the necessary representational capacity for training on AMASS. All ablations and \emph{MVAE} are trained in the exact same way as the full model. Additionally, when used in test-time optimization we use the same energy weightings as described in~\cref{table:energysettings} but with irrelevant energies removed (\eg the \emph{No Contacts} ablation does not allow the use of $\en{env}$). Note that $\en{init}$ \emph{is still used} with \emph{MVAE} and all ablations, the only thing that changes is the prior in $\en{CVAE}$.

\paragraph{Motion Estimation Baselines}
The \emph{VPoser-t} baseline is exactly the initialization phase of our proposed test-time optimization, \ie we use weightings in \cref{table:energysettings}.

The \emph{PROX-RGB} baseline fits the neutral SMPL-X~\cite{pavlakos2019smplx} body model to the same 2D OpenPose detections used by our method. It does not use the face or hand keypoints for fitting similar to our approach. The \emph{PROX-D} baseline uses the fittings provided with the PROX dataset, which are on the known gendered SMPL-X body model and use face/hand 2D keypoints for fitting. 

The \emph{VIBE} baseline uses the same 2D OpenPose detections as our method in order to define bounding boxes for inference. We found this makes for a more fair comparison since the real-time trackers used in their implementation\footnote{\href{https://github.com/mkocabas/VIBE}{see Github}} often fail for medium to heavy occlusions common in our evaluation datasets.

\paragraph{Evaluation Metrics}
In order to report occluded (\textbf{Occ}) and visible (\textbf{Vis}) positional errors separately, we must determine which joints/keypoints are occluded during evaluation. This is easily done for 3D tasks where ``occlusions'' are synthetically generated. For RGB data in i3DB, we use the person segmentation mask obtained with DeepLabv3~\cite{chen2017rethinking} to determine if a ground truth 3D joint is visible after projecting it to the camera.

For a joint $\pnt_t \in \smpljoints_t$ at time $t$ the acceleration magnitude (\textbf{Accel}) is computed as
\begin{equation}
    a = || (\pnt_{t-1} - 2\pnt_t + \pnt_{t+1}) / h^2 ||
\end{equation}
where $h = 1/ 30$ for all datasets. Ground penetration frequency (\textbf{Freq}) for a given penetration threshold $g_\text{thresh}$ is computed over all $D$ frames in a dataset as
\begin{equation}
    \dfrac{\sum_D \mathbf{1}(d_\text{pen}^\text{ltoe} > g_\text{thresh} ) + \mathbf{1}(d_\text{pen}^\text{rtoe} > g_\text{thresh} )}{2D}
\end{equation}
where $\mathbf{1}(\cdot)$ is the indicator function and $d_\text{pen}^\text{ltoe}, d_\text{pen}^\text{rtoe}$ are the penetration distances (shortest distance to the ground plane) for the left and right toe joints at the current frame.

\subsection{Estimation from 3D Observations}
For fitting to 3D data, as presented in Sec.~5.4 of the main paper, the observation and canonical coordinate frames are identical since AMASS data is used, therefore \optname does not optimize the ground plane $\floor$.

\subsection{Estimation from RGB(-D) Observations}
\paragraph{i3DB} Positional joint errors are computed using a 12-joint subset of the ground truth 3D joint annotations which correspond to the SMPL joints used by our method and all baselines. These include ankles, knees, wrists, elbows, shoulders, the neck, and the root. The leg joints reported in Tab. 3 of the main paper include the ankles and knees.

In fitting to i3DB videos, we mask 2D joint observations using the person segmentation mask which we found beneficial under the numerous severe occlusions where OpenPose may predict spurious 2D pose with incorrectly high confidence.

\paragraph{PROX} For fitting to PROX RGB(-D) videos we found it best \emph{not} to mask out 2D joints based on the person segmentation mask, as occlusions are typically minor and OpenPose is relatively accurate. However, the segmentation is always used on the point cloud back-projected from the depth map to ignore points far from the person.

\section{Extended Evaluations}
\label{supp:sec:evals}
In this section, we present experimental evaluations to supplement those in Sec. 5 of the main paper, which were ommitted due to space constraints.

\begin{table*}
\setlength{\tabcolsep}{7pt}
\begin{center}
\begin{tabular}{l | c c c c | c c c c | c c c}
 & \multicolumn{4}{c|}{Global Joint Error}  & \multicolumn{4}{c|}{Root-Aligned Joint Error}  & & \multicolumn{2}{c}{Ground Pen} \\
\textbf{Method} & \textbf{Vis} & \textbf{Occ} & \textbf{All} & \textbf{Legs} & \textbf{Vis} & \textbf{Occ} & \textbf{All} &  \textbf{Legs} & \textbf{Accel} & \textbf{Freq} & \textbf{Dist} \\
\hline
No $\en{init}$ & 28.04 & 39.37 & 30.96 & 34.91 & 12.11 & 22.27 & 14.73 & 21.40 & 2.57 & 2.80\% & 0.85 \\
No $\en{c}$ & 29.17	& 40.62 & 32.12 & 35.83 & 12.99 & 24.34 & 15.92 & 22.66 & 2.75 & \textbf{1.98\%} & 0.67 \\
No $\en{b}$ & 28.29 & 39.77 & 31.25 & 34.37 & 13.00 & 24.36 & 15.93 & 22.31 & 2.93 & 4.43\% & 1.13 \\
No $\en{skel}$  & 32.87 & 44.16 & 35.78 & 37.55 & 14.67 & 26.82 & 17.80 & 24.18 & 3.81 & 4.35\% & 1.29 \\
No $\en{env}$ & 26.83 & 37.62 & 29.61 & 32.91 & 12.09 & 22.04 & 14.66 & 20.96 & 2.51 & 2.10\% & \textbf{0.66} \\
\hline
Full Energy &  \textbf{26.00} & \textbf{34.36} & \textbf{28.15} & \textbf{31.26} & \textbf{12.02} & \textbf{21.70} & \textbf{14.51} & \textbf{20.74} & \textbf{2.43} & 2.12\% & 0.68 
\end{tabular}
\end{center}
\caption{Motion and shape from RGB video (\ie 2D joints) on i3DB~\cite{monszpart2019iMapper}. Joint errors are in $cm$ and acceleration is $m/s^2$. Results use \optname with \methodname. The top part shows various energy terms ablated.}
\label{table:energyablation}
\end{table*}

\subsection{Qualitative Evaluation}
Please see the \textbf{supplementary videos} for extensive qualitative results corresponding to each experiment in Sec. 5 of the main paper. In this document, we show various representative examples from these videos and summarize important results.

\begin{figure*}
\begin{center}
\includegraphics[width=\linewidth]{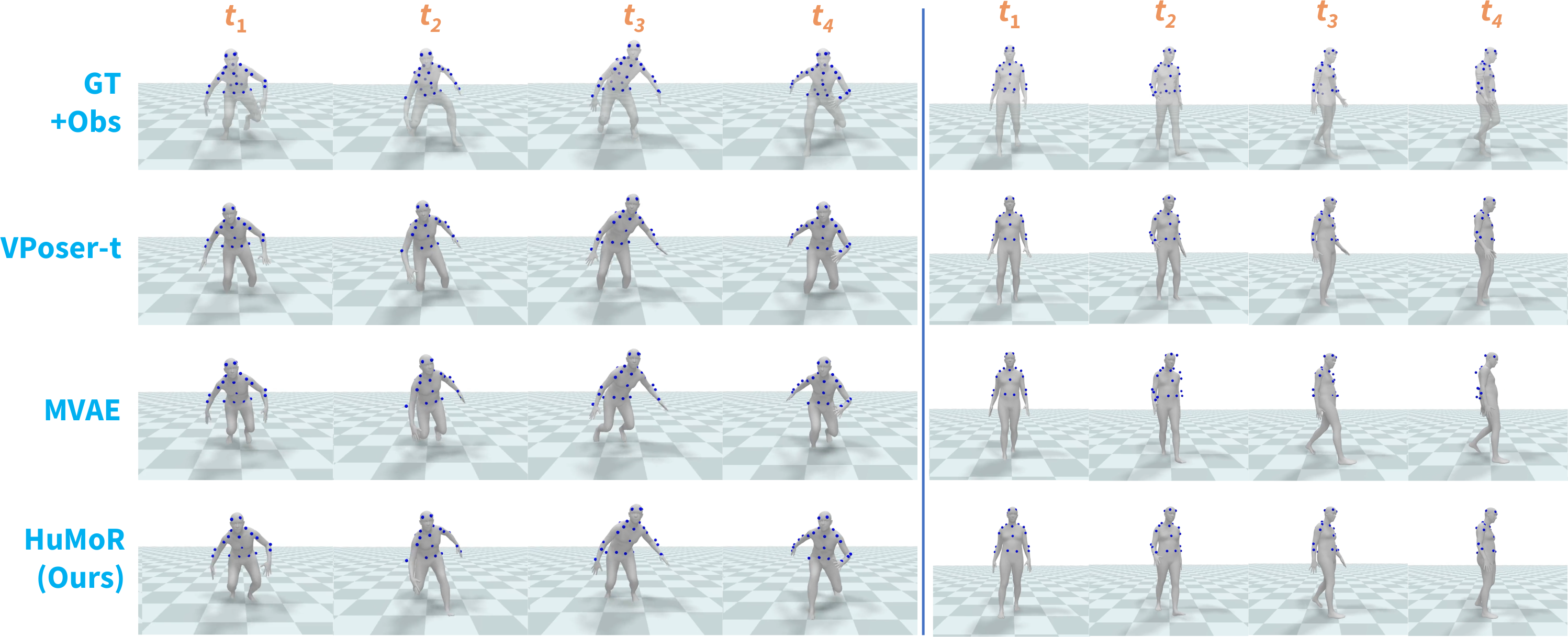}
\end{center}
\captionof{figure}{Comparison to baselines when fitting to 3D keypoints from held out sequences in the AMASS dataset. \emph{GT+Obs} shows the ground truth body motion and observed keypoints in blue, while each method output shows the predicted motion with the observed keypoints for reference.\vspace{-1mm}}
\label{fig:qual-keypts3d}
\end{figure*}

\cref{fig:qual-keypts3d} shows results fitting to occluded 3D keypoints (Sec. 5.4 of the main paper). Performance of \optname with \methodname is compared to the \emph{VPoser-t} and \emph{MVAE} baselines on two sequences. \emph{VPoser-t} fails to produce any plausible lower-body motion since it uses only a pose prior, while using \emph{MVAE} as the motion prior often gives unnatural and implausible motions that don't align well with the observed keypoints.

\begin{figure*}
\begin{center}
\includegraphics[width=\linewidth]{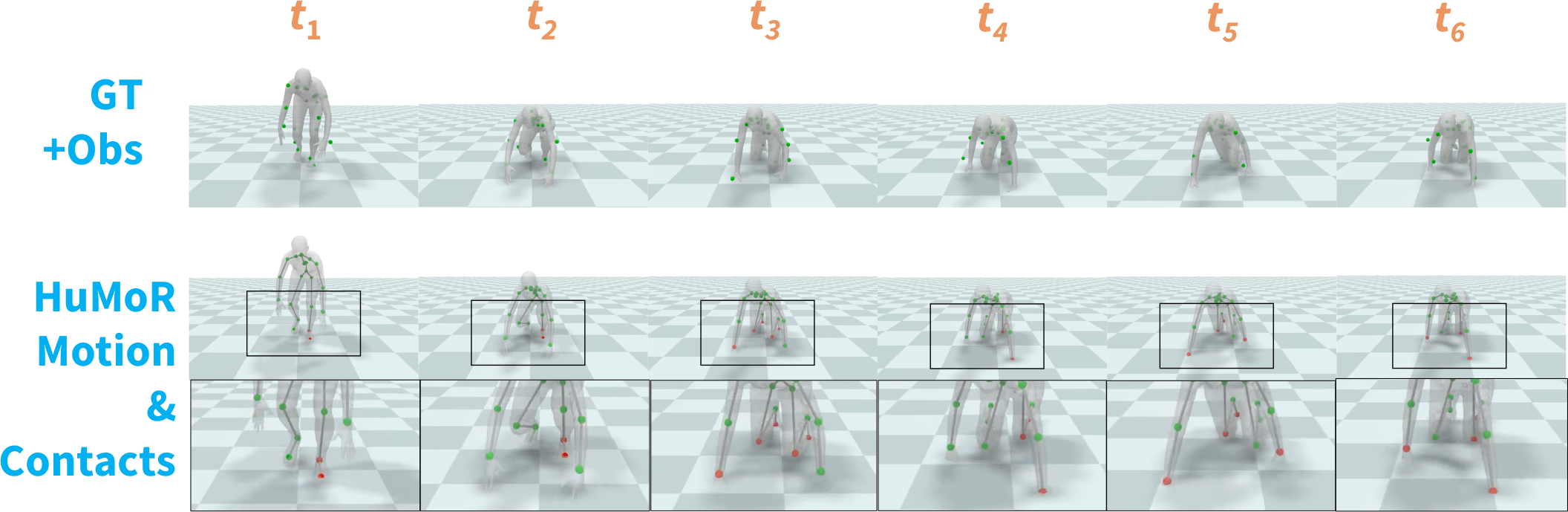}
\end{center}
\captionof{figure}{\optname with \methodname recovers complex contact patterns involving feet, knees, and hands from noisy 3D joint observations (shown on top along with the ground truth motion).} %
\label{fig:qual-joints3d}
\end{figure*}

\cref{fig:qual-joints3d} shows results using \optname with \methodname for fitting to noisy 3D joints (Sec. 5.4 of the main paper). Both the estimated motion and contacts are shown for a crawling sequence. Note that HuMoR recovers complex contact patterns involving not only the feet, but also hands and knees.

\begin{figure*}
\begin{center}
\includegraphics[width=\linewidth]{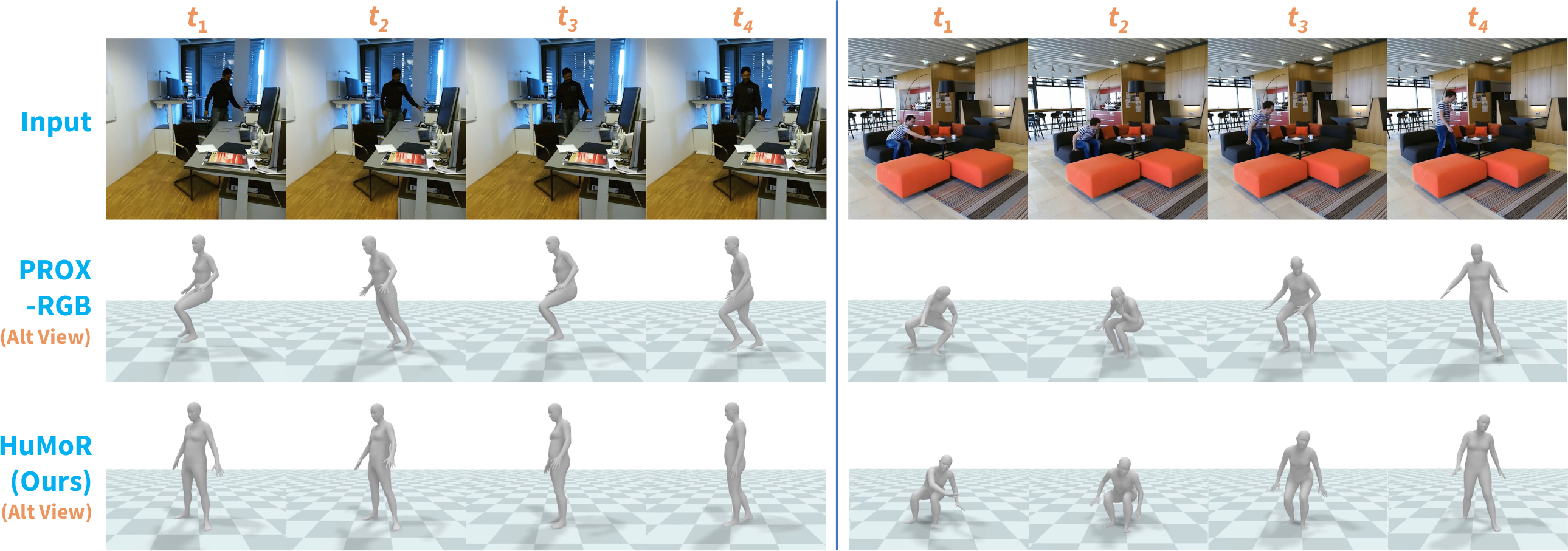}
\end{center}
\captionof{figure}{Comparison to \emph{PROX-RGB} on videos from the PROX dataset. Predicted motion is shown from an alternate viewpoint for both methods. In both sequences, \emph{PROX-RGB} is temporally inconsistent. In the left example, the lower body occlusion causes implausible neutral standing or sitting poses. The predicted ground plane from \methodname is shown for reference.} %
\label{fig:qual-prox-rgb}
\end{figure*}

\cref{fig:qual-prox-rgb} demonstrates fitting performance on RGB videos from PROX compared to the \emph{PROX-RGB} baseline (Sec. 5.5 of the main paper). \emph{PROX-RGB} produces temporally incoherent results since it operates on single frames. However, it also uses the scene mesh as input which allows for plausible poses when the person is fully visible. This does not greatly improve results under occlusions, though, often reverting to a mean leg pose similar to \emph{VPoser-t} and \emph{VIBE}. 

\begin{figure*}
\begin{center}
\includegraphics[width=\linewidth]{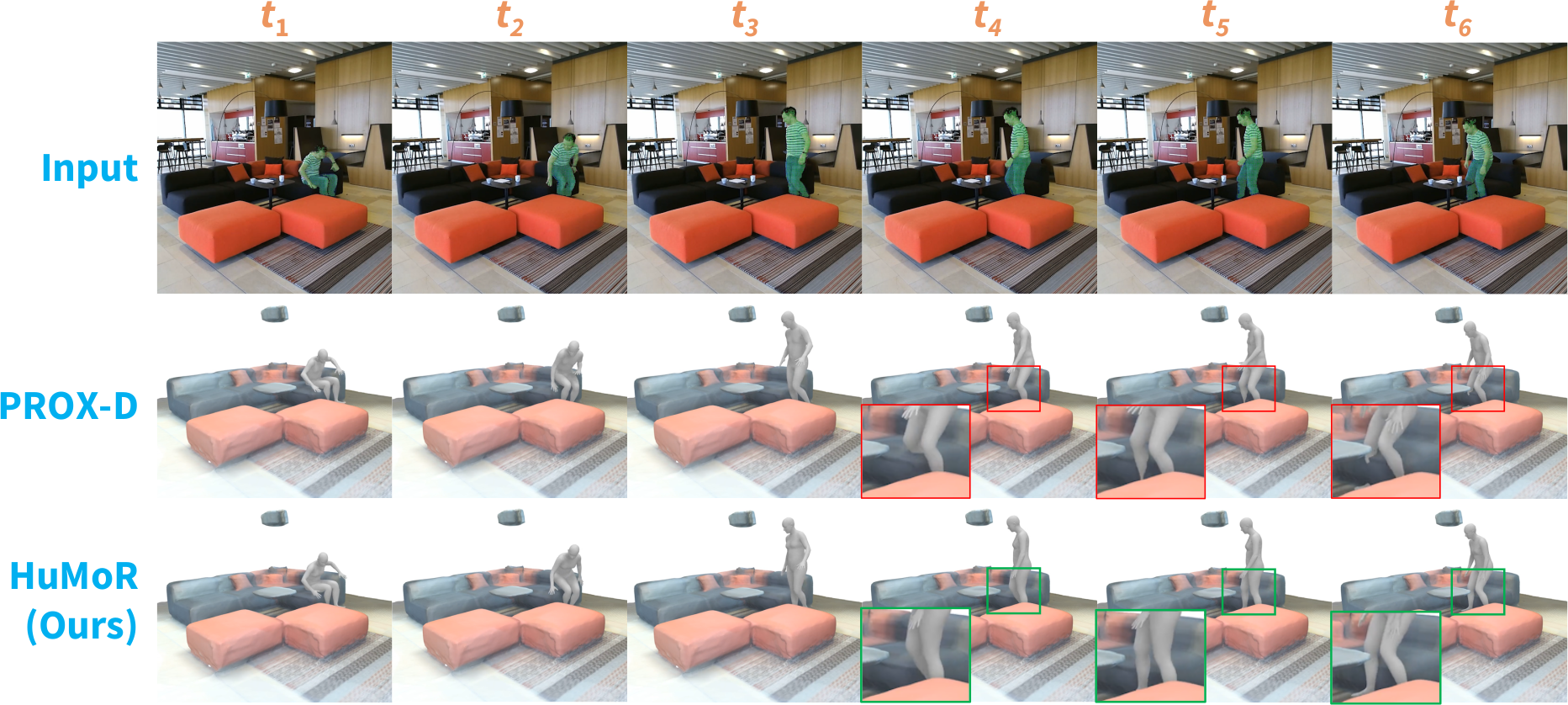}
\end{center}
\captionof{figure}{Comparison to \emph{PROX-D} on PROX RGB-D data. Motion and shape results are rendered within the ground truth scene mesh for reference. Though our method does not use the scene mesh as input like \emph{PROX-D}, it still produces motions that are plausible within the environment by using HuMoR as the motion prior, sometimes better than \emph{PROX-D} as indicated by the red and green boxes here.} %
\label{fig:qual-prox-rgbd}
\end{figure*}

\cref{fig:qual-prox-rgbd} demonstrates fitting to RGB-D videos from PROX compared to the \emph{PROX-D} baseline (Sec. 5.5 of the main paper). Using motion as a prior allows for natural interaction within the scene, as detailed in the figure caption.

\begin{figure*}
\begin{center}
\includegraphics[width=\linewidth]{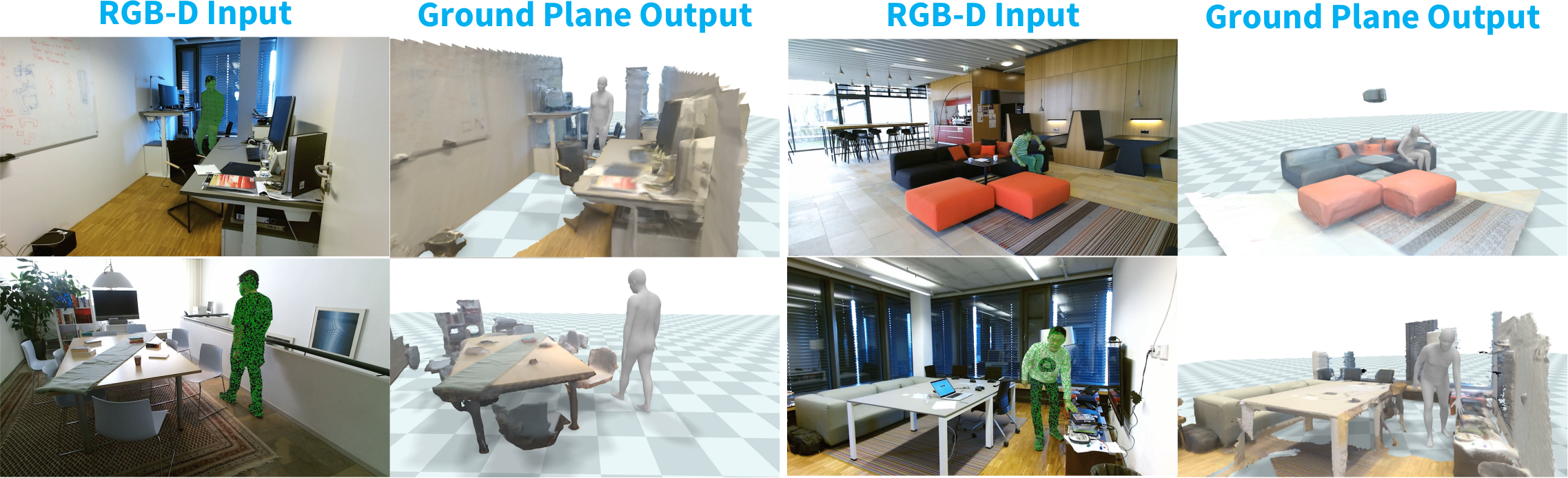}
\end{center}
\captionof{figure}{Ground estimation examples from each scene in PROX from RGB-D. The scene mesh is shown for reference only, it is not an input or output of \optname with \methodname.\vspace{-1mm}}
\label{fig:qual-ground}
\end{figure*}

\cref{fig:qual-ground} shows ground plane estimations when fitting to RGB-D data for each of the scenes in our PROX dataset. The estimated floor is rendered within the true scene mesh for reference.

\begin{figure}[t]
\begin{center}
  \includegraphics[width=0.95\linewidth]{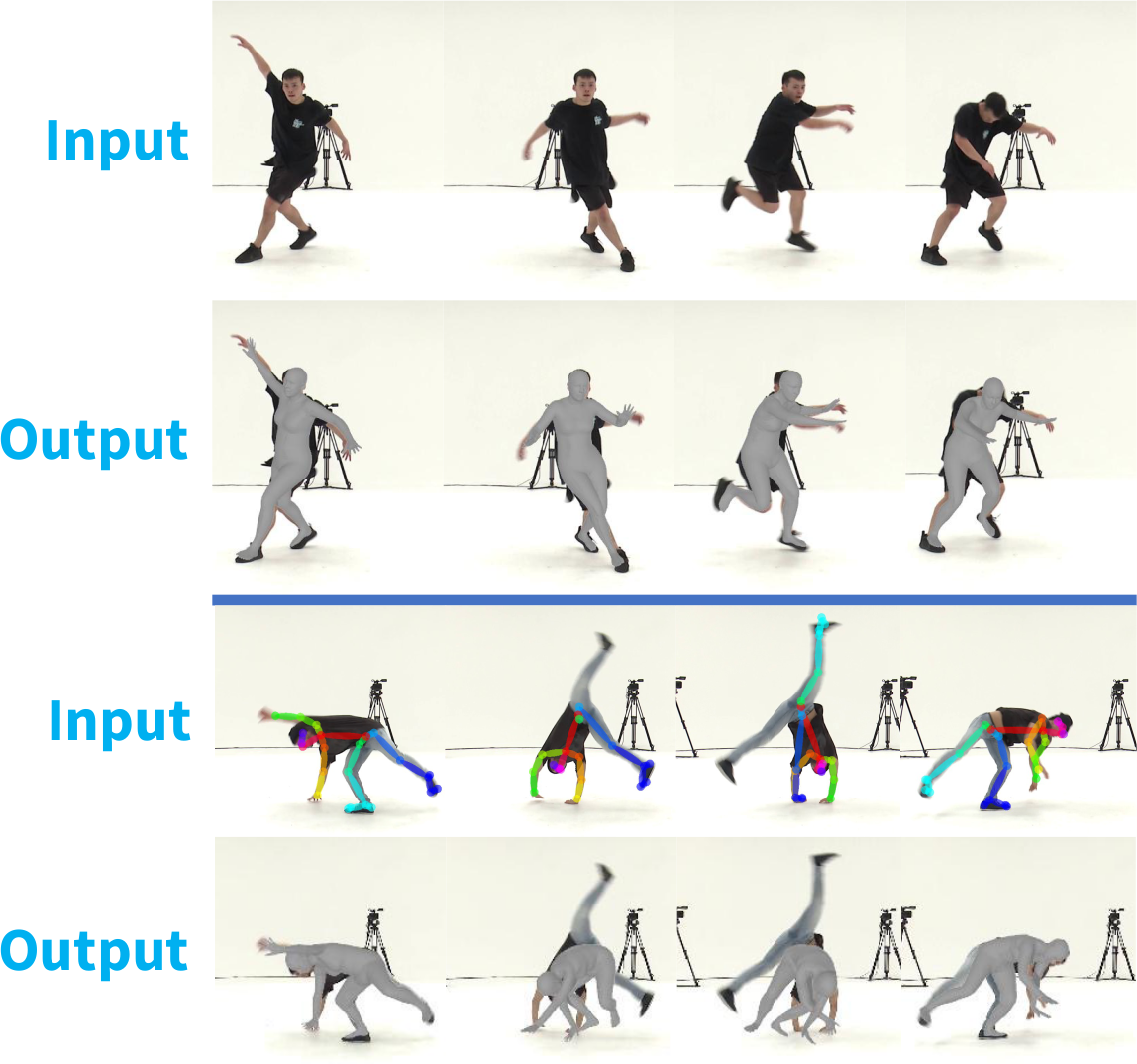}
\end{center}
\captionof{figure}{Example sequences using TestOpt with HuMoR to fit to 2D joints in AIST dance videos~\cite{TsuchidaAist2019}. HuMoR generalizes to these highly dynamic motions and robustly recovers from inaccurate 2D joint detections (bottom).}
\label{fig:qual-dance}
\end{figure}

\begin{figure*}[t]
\begin{center}
  \includegraphics[width=0.95\linewidth]{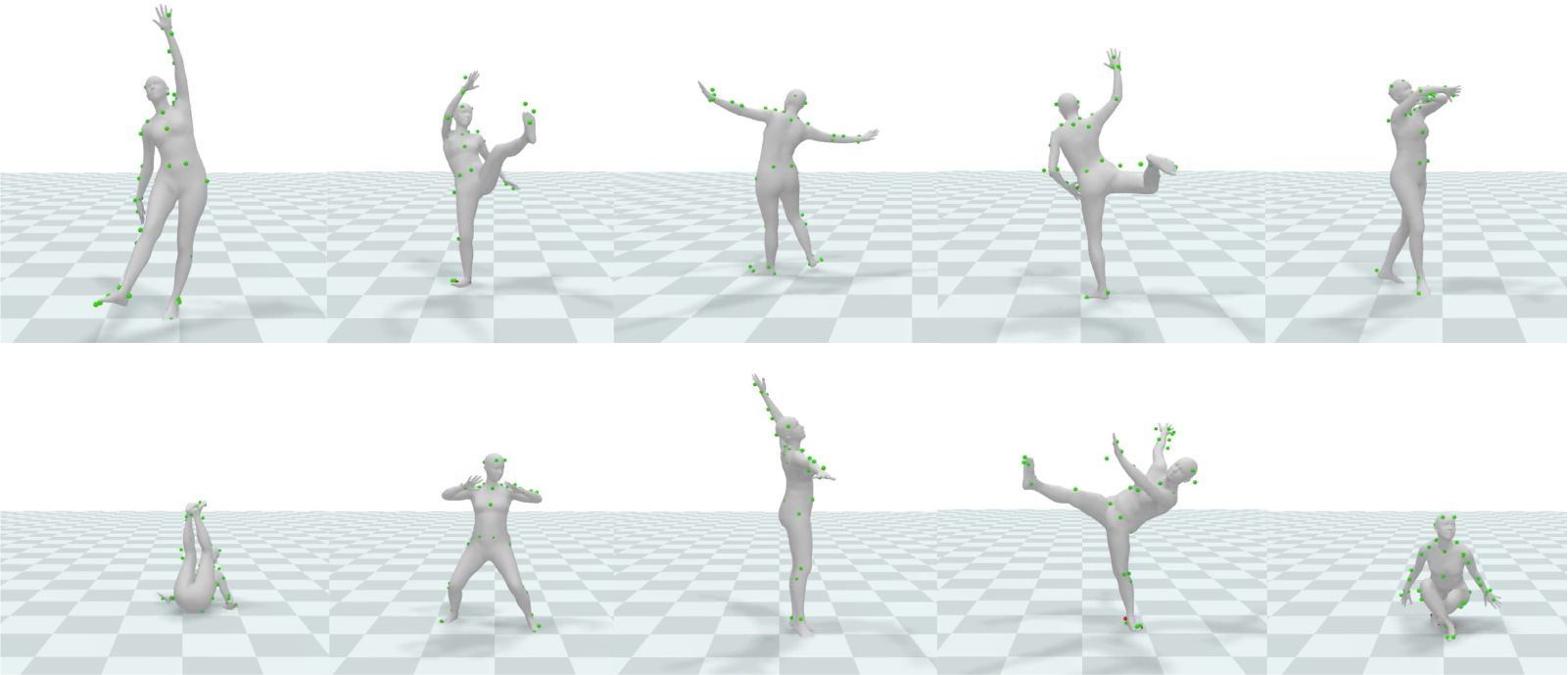}
\end{center}
\captionof{figure}{Example frames from using TestOpt with HuMoR to fit to full-body 3D keypoint motions from dynamic DanceDB~\cite{AristidouDanceDB2019} data in AMASS~\cite{AMASS:2019}. Estimated body shape and pose is shown along with ground truth keypoints in green. Despite not training on this dance data, HuMoR is able to effectively generalize to capture these complex motions.}
\label{fig:qual-dance3d}
\end{figure*}

Finally, we evaluate TestOpt with HuMoR on highly dynamic dancing data to demonstrate the generalization ability of the CVAE motion model. \cref{fig:qual-dance3d} shows a sample of frames from motions fit to the DanceDB \cite{AristidouDanceDB2019} subset of AMASS~\cite{AMASS:2019}. In this case, the observations are full-body 3D keypoints. Though HuMoR is trained on data with few dancing motions, it is able to capture these difficult motions at test time since it only operates on pairs of frames. Additionally, \cref{fig:qual-dance} shows fitting results on RGB videos from the AIST dance dataset~\cite{TsuchidaAist2019}. Since HuMoR allows for large accelerations, it accurately generalizes to fast motions (top - note motion blur). Moreover, it is able to recover from poor 2D joint detections from OpenPose due to the cartwheel motion (bottom).

\subsection{Optimization Objective Ablation}
In this experiment, we analyze the effect of the energy terms and regularizers in our test-time optimization (\optname) formulation. \cref{table:energyablation} reports results on the i3DB~\cite{monszpart2019iMapper} dataset using \optname with \methodname for different energy ablations.

\emph{No $\en{init}$} does not use the initial state Gaussian mixture model (GMM) as part of the motion prior (\ie assumes a uniform prior over the initial state). This means the input $\xzero$ to CVAE rollout may not be plausible, especially early in optimization, leading to degraded performance. \emph{No $\en{c}$} and \emph{No $\en{b}$} remove individual terms of the skeleton regularization: the joint and bone length consistency. \emph{No $\en{skel}$} removes the entire skeleton regularization (both $\en{c}$ and $\en{b}$), severely affecting final performance. This term is important to ensuring the CVAE is actually rolling out realistic motions. Finally, \emph{No $\en{env}$} removes the contact velocity and height terms, increasing errors particularly for occluded joints while resulting in similar plausibility metrics. 

\subsection{Sensitivity to Occlusions and Noise}
Next, we look at performance on estimation from 3D data under increasing levels of occlusions and noise. Similar to Sec. 5.4 in the main paper, we consider fitting to occluded 3D keypoints (points under a given height threshold are unobserved) and 3D joint locations with added Gaussian noise. For this experiment, we use the held out Transitions subset of AMASS~\cite{mahmood2019amass}.

\begin{table}
\setlength{\tabcolsep}{5pt}
\begin{center}
\begin{tabular}{l | c c}
\textbf{Dataset} & \textbf{Batch Size} & \textbf{Mean Seq Time} \\
\hline
AMASS (occ keypoints) & 12 & 2.95 \\
AMASS (noisy joints) & 12 & 2.45 \\
i3DB (RGB) &  6 & 6.42 \\
PROX (RGB) & 6 & 4.48 \\
PROX (RGB-D) & 6 & 4.85 \\
\end{tabular}
\end{center}
\caption{Mean per-sequence optimization times (in minutes) for evaluations on each dataset. Optimizations are done on batches of $3s$ (90 frame) sequences.\vspace{-3mm}}
\label{table:computation}
\end{table}

\begin{figure*}
\begin{center}
\includegraphics[width=\linewidth]{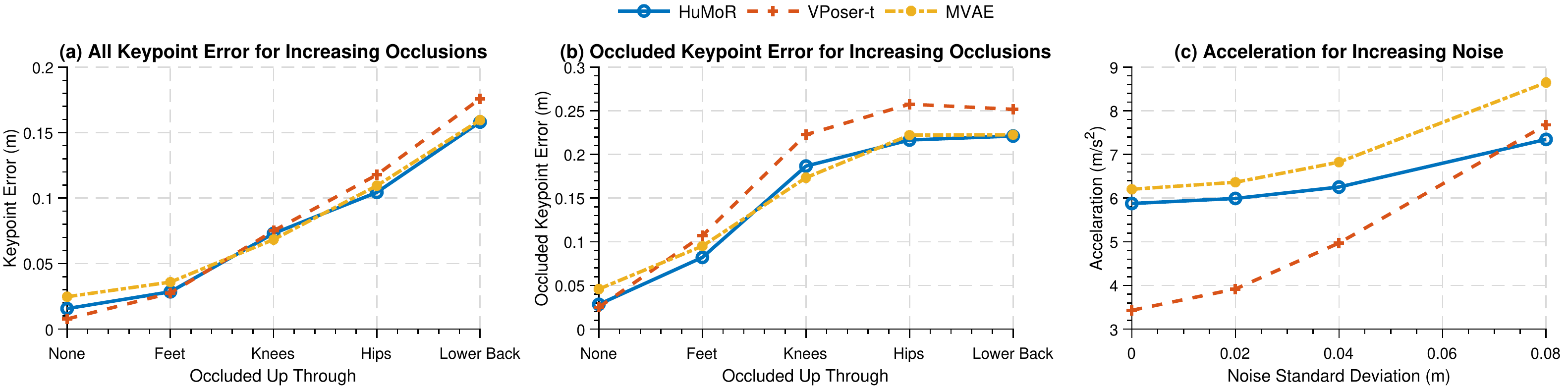}
\end{center}
\captionof{figure}{Keypoint errors for \textbf{all} (\textbf{a}) and \textbf{occluded} (\textbf{b}) points for increasing levels of occlusion when fitting to 3D keypoints with \optname. (\textbf{c}) Joint acceleration magnitude for increasing levels of noise when fitting to 3D joints with \optname.} %
\label{fig:sens-plots}
\end{figure*}

Mean keypoint errors for \textbf{all} and \textbf{occluded} points are shown for increasing occlusions in \cref{fig:sens-plots}(a)(b). From left to right the occluded height threshold is 0.0, 0.3, 0.6, 0.9, and 1.2 $m$ which roughly corresponds to the lower body being occluded from the floor up through the body parts on the $x$-axis. With no occlusions (\emph{None}) \emph{VPoser-t} most closely fits the clean points, while \methodname outperforms \textit{MVAE} due to improved expressiveness. As occlusions increase, \methodname and \textit{MVAE} perform similarly while occluded errors increase greatly for \emph{VPoser-t} as also observed in other experiments. Note that after the knees become occluded, errors for occluded keypoints tend to saturate as performance is dependent nearly entirely on the motion prior.

Mean joint acceleration magnitude is shown for increasingly noisy 3D joint observations in \cref{fig:sens-plots}(c). From left to right noise increases to 8 $cm$ standard deviation. Importantly, the performance of \methodname stays relatively stable. \methodname increases only 25\% while \emph{VPoser-t} and \textit{MVAE} increase 124\% and 57\%, respectively.

\subsection{Computational Requirements of \optname}
For all experiments presented in the main paper, we perform optimization on batches of $3s$ sequences (90 frames). \cref{table:computation} shows the mean per-sequence optimization times for each of these experiments where \emph{AMASS} corresponds to experiments in Sec. 5.4 of the paper and \emph{i3DB} and \emph{PROX} correspond to Sec. 5.5. The per-sequence time is computed by taking the total time to optimize over the whole dataset divided by the number of $3s$ sequences. Note that batching speeds up this large scale optimization significantly, \ie optimizing a single sequence will only be slightly faster than a batch of sequences since the primary bottleneck is CVAE rollout. All batched optimizations were performed on a 24 GB Titan RTX GPU, though an 8 GB GPU is sufficient to optimize a single sequence.


\end{document}